\documentclass[10pt,journal,cspaper,compsoc]{IEEEtran}

\usepackage{epsfig}
\usepackage{graphicx}
\usepackage{amsmath}
\usepackage{amssymb}
\usepackage{bm}
\usepackage{multirow}
\usepackage{array}
\usepackage{paralist}
\usepackage{xspace}
\usepackage{booktabs}
\usepackage[pagebackref=true,breaklinks=true,colorlinks,linkcolor=blue,citecolor=blue,bookmarks=true]{hyperref}

\newcommand{\bbR}{\mathbb{R}}

\newcommand{\bI}{\mathbf{I}}
\newcommand{\by}{\mathbf{y}}
\newcommand{\bt}{\mathbf{t}}
\newcommand{\bp}{\mathbf{p}}
\newcommand{\bB}{\mathbf{B}}

\newcommand{\bz}{\mathbf{z}}
\newcommand{\hB}{\hat{B}}
\newcommand{\hb}{\hat{b}}

\def\etal{et~al.\xspace}

\begin{document}

\title{Progressive Representation Adaptation for Weakly Supervised Object Localization}

\author{
Dong~Li, Jia-Bin~Huang, Yali~Li, Shengjin~Wang$^{\star}$ and Ming-Hsuan~Yang
\IEEEcompsocitemizethanks{
\IEEEcompsocthanksitem Dong Li is with the Department of Electronic Engineering, 
Tsinghua University, Beijing, China. E-mail: lidong12@tsinghua.org.cn 
\IEEEcompsocthanksitem Jia-Bin Huang is with the Department of Electrical and Computer Engineering, Virginia Tech, Blacksburg, VA, 24060. E-mail: jbhuang@vt.edu 
\IEEEcompsocthanksitem Yali Li is with the Department of Electronic Engineering, Tsinghua University, Beijing, China. E-mail: liyali@ocrserv.ee.tsinghua.edu.cn 
\IEEEcompsocthanksitem Shengjin Wang is with the Department of Electronic Engineering, 
Tsinghua University, Beijing, China. E-mail: wgsgj@tsinghua.edu.cn 
\IEEEcompsocthanksitem Ming-Hsuan Yang is with the School of Engineering, University of California, Merced, CA, 95344. E-mail: mhyang@ucmerced.edu
}
\thanks{$^{\star}$Corresponding author.}}

\IEEEcompsoctitleabstractindextext{
\begin{abstract}
We address the problem of weakly supervised object localization where only image-level annotations are available for training object detectors.
Numerous methods have been proposed to tackle this problem through mining object proposals.
However, a substantial amount of noise in object proposals causes ambiguities for learning discriminative object models.
Such approaches are sensitive to model initialization and often converge to undesirable local minimum solutions. 
In this paper, we propose to overcome these drawbacks by progressive representation adaptation with two main steps: 1) classification adaptation and 2) detection adaptation.
In classification adaptation, we transfer a pre-trained network to a multi-label classification task for recognizing the presence of a certain object in an image.
Through the classification adaptation step, the network learns discriminative representations that are specific to object categories of interest.
In detection adaptation, we mine class-specific object proposals by exploiting two scoring strategies based on the adapted classification network.
Class-specific proposal mining helps remove substantial noise from the background clutter and potential confusion from similar objects.
We further refine these proposals using multiple instance learning and segmentation cues. 
Using these refined object bounding boxes, we fine-tune all the layer of the classification network and obtain a fully adapted detection network.
We present detailed experimental validation on the PASCAL VOC and ILSVRC datasets. 
Experimental results demonstrate that our progressive representation adaptation algorithm performs favorably against the state-of-the-art methods.
\end{abstract}

}

\maketitle

\IEEEdisplaynotcompsoctitleabstractindextext

\IEEEpeerreviewmaketitle

\section{Introduction}

\begin{figure}[t]
\centering
\includegraphics[width=\linewidth]{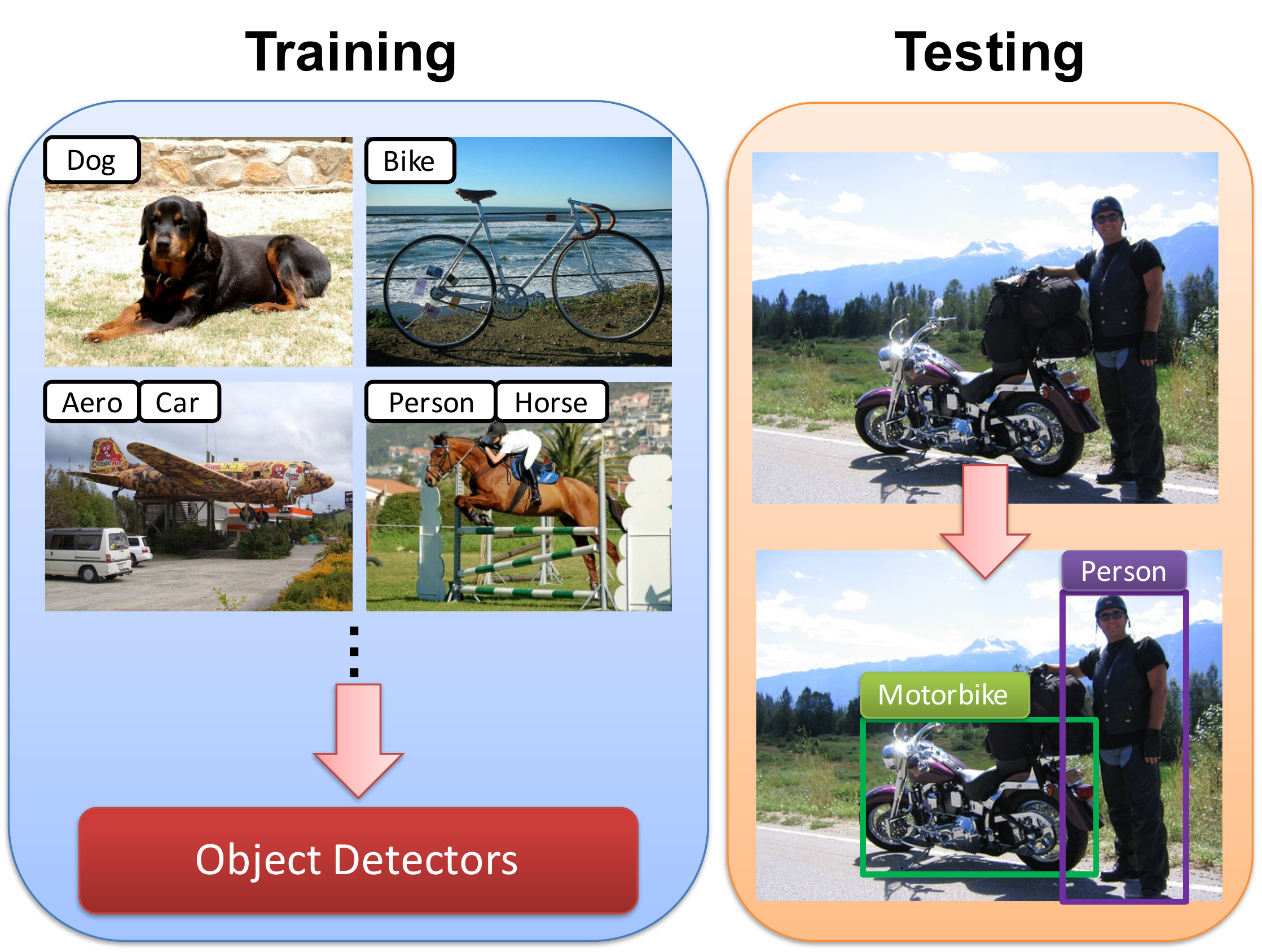}
\caption{\textbf{Weakly supervised object localization}. Given a collection of training images with image-level annotations, our goal is to train object detectors for localizing objects in unseen images.}
\label{figure:setting}
\end{figure}

\IEEEPARstart{O}{bject} localization is a fundamental building block for image understanding. 
It aims to identify all instances of particular object categories (e.g., person, cat, and car) in images.
The main challenges in object localization lie in constructing object appearance models for handling large intra-class variations and complex background clutters. 
The state-of-the-art approaches typically train object detectors from a large and diverse set of training images~\cite{ren2015faster, redmon2016yolo9000} in a fully supervised manner.
However, such a fully supervised learning paradigm relies on \emph{instance-level} annotations, e.g., tight bounding boxes, which are time-consuming and labor-intensive. 
In this paper, we focus on the \emph{weakly supervised} object localization problem where only image-level labels indicating the presence or absence of an object category are available for training. 
Figure~\ref{figure:setting} illustrates the problem setting. 
This particular setting is important for large-scale practical applications because image-level annotations are often readily available from the Internet, e.g., through text tags~\cite{guillaumin2009tagprop}, GPS tags~\cite{doersch2012makes}, and image search queries~\cite{li2013harvesting}.

Weakly supervised learning (WSL)~\cite{li2016weakly} is a challenging problem primarily due to the large gap between a source domain (weakly annotated data) and the corresponding target task (object detection).
Most existing methods~\cite{song2014learning,cinbis2017weakly,bilen2015weakly,bilen2014weakly,siva2012defence,siva2011weakly,song2014weakly,siva2013looking} formulate WSL as a multiple instance learning (MIL) problem. 
In the MIL framework, each image consists of a bag of potential object instances.
Positive images are assumed to contain \emph{at least} one object instance of a certain object category and negative images do not contain object instances from this category. 
Using this weak supervisory signal, WSL methods often alternate between 
(1) selecting the positive object instances from positive images and 
(2) learning object detectors. 
However, due to the non-convexity, these methods are sensitive to model initialization and prone to getting trapped into local extrema. 
Although many efforts have been made to alleviate the problem via seeking better initialization models~\cite{song2014learning,song2014weakly,siva2013looking,siva2011weakly,siva2012defence} and optimization strategies~\cite{cinbis2017weakly,bilen2015weakly,bilen2014weakly}, the quality of object instance selection is still limited. 
We observe that previous MIL based methods attempt to train object detectors directly from a large and noisy collection of object proposals. 
The noise in a collection of proposals makes learning discriminative object-level features challenging and may lead
to inaccurate localization for training the respective detector.

Recent methods also leverage transfer learning and domain adaptation techniques to address the WSL problem~\cite{rochan2015weakly,hoffman2014lsda,hoffman2014detector,shi2012transfer,shi2017weakly,kumar2016track}.
These approaches often require additional annotated data (bounding box annotations) or pre-trained detectors for several object categories.
Examples of transferred knowledge include similar appearance~\cite{rochan2015weakly, shi2017weakly}, spatial information from visual tracking~\cite{kumar2016track} and object size prior~\cite{shi2016weakly}. 

While existing schemes have shown promising results, three drawbacks remain to be addressed. 
First, it is difficult to select correct object proposals because the collection of category-independent proposals contains many noisy results (e.g., background clutter, object parts).
Typically, only a few out of several thousands of proposals are actual object instances.
Second, some approaches use pre-trained CNNs as feature extractors and do not adapt the weights from whole-image classification to object detection.
Third, domain adaptation based methods often require either auxiliary strongly annotated data or pre-trained detectors.

\begin{figure}[t]
\centering
\includegraphics[width=\linewidth]{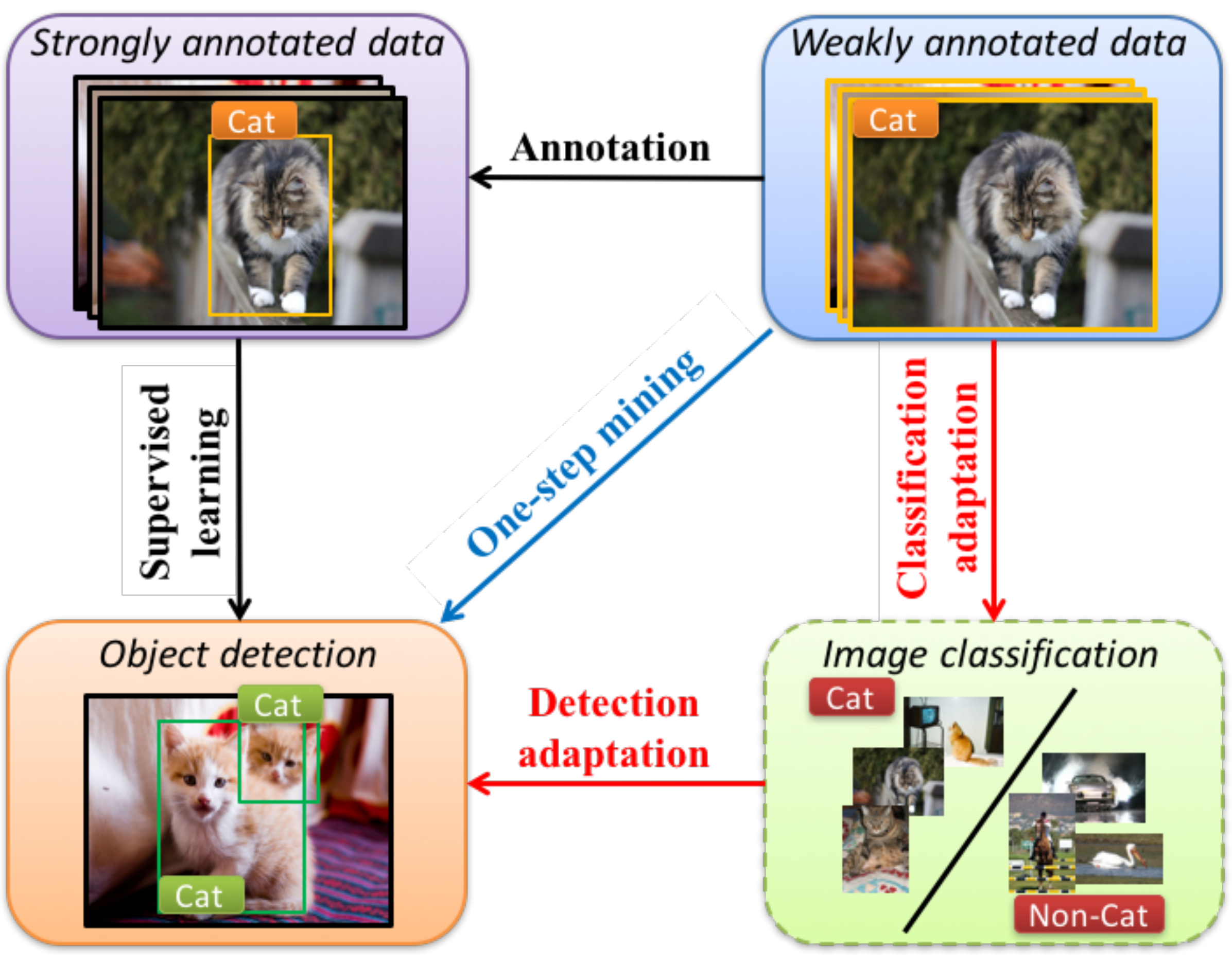} 
\caption{\textbf{Progressive adaptation.} 
Strongly supervised methods use instance-level annotations (e.g., tight bounding boxes) to train object detectors.
Most weakly supervised methods use mine object proposals to select object instances from a large and noisy candidate pool in one single step.
We propose a two-step progressive adaptation approach: classification adaptation (Section~\ref{section:cls-adapt}) and detection adaptation (Section~\ref{section:det-adapt}). 
Our approach effectively filter out the noisy object proposal collection and thus can mine confident candidates for learning discriminative object detectors.
}
\label{figure:difference}
\end{figure}

In this paper, we propose a two-step domain adaptation algorithm for weakly supervised object localization based on 1) classification adaptation and 2) detection adaptation. 
Figure~\ref{figure:difference} illustrates the major difference between the proposed algorithm and existing work. 
Our key observation is that it is difficult to train object detectors directly under weak supervisory signals due to the substantial amount of noise in the object proposal collections. 
Essentially, the main difficulty arises from the large gap between the source domain and target task, as shown in the top-right and bottom-left corner of Figure~\ref{figure:difference}. 
The goal of our work is to bridge the gap through progressive representation adaptation. 
In the classification adaptation step, we train a classification network using the given weak image-level labels.
The classification network can recognize the presence of a certain object category in an image.
While the classification network is trained for image-level classification, the network provides discriminative representation for localizing target objects.
In the detection adaptation step, we first use the classification network to collect class-specific object proposals.
To this end, we explore two strategies for scoring the collection of object proposals.
Our first scoring strategy computes the confidence differences between the candidate proposal and its mask-out image based on the outputs of the classification network.
Our second approach localizes the discriminative image regions using class-specific feature maps.
By combining the two proposal scoring strategies to select class-specific object proposals, we significantly alleviate the negative effect of background clutters and potential confusion from other categories.
We then apply a multiple instance learning algorithm to mine confident candidates and refine them using segmentation cues. 
The derived tight object bounding boxes are used to fine-tune all layers, thereby adapting the classification network to a detection network.

The proposed algorithm addresses the drawbacks of prior work in the following three aspects: 
(1) The classification adaptation step fine-tunes the pre-trained network from the 
ImageNet such that it can collect \emph{class-specific} object proposals with higher precision. 
This step aims at removing object proposals that correspond to background clutters and potential confusion from similar object categories, leading to a \emph{purified} collection of object candidates for the subsequent multiple instance learning algorithm. 
(2) The detection adaptation uses confident object candidates to optimize the CNN representations for the target domain. 
This step aims at adapting the image classification network into object detectors, providing more discriminative feature representations for localizing generic objects (instead of the presence of them) in an image. 
(3) Our method learns object detectors from weakly annotated data \emph{without} any strong labels (e.g., bounding box annotations).

We make the following contributions in this work:
\begin{compactenum}
\item We propose to address the weakly supervised object localization problem by \emph{progressive} representation adaptation. 
In contrast to most existing methods which directly train detectors from a large set of 
noisy object proposals, we select high-quality object candidates and learn discriminative representations for object detection. 
Progressive learning decomposes the challenging problem of domain adaptation between source domain (image-level annotated data) and target task (instance-level object detection) into two relatively simpler tasks. 
Our results show that this strategy of progressive learning is crucial for good performance.
\item Class-independent region proposals (e.g.,~\cite{zitnick2014edge}) are widely used in modern object detection frameworks~\cite{girshick2015fast,ren2015faster}. 
In the setting of WSL, we instead collect \emph{class-specific} object proposals to alleviate the impact of background clutter and confusion from other categories. 
To this end, we explore two scoring strategies, i.e., contrast and activation scores. 
The contrast scores are derived from the classification layer of the adapted classification network. 
The activation scores, on the other hand, are derived from the last convolutional feature maps. 
Our results show that combination of these two scores improves the quality of the mined object proposals.
\item We observe that tight object bounding boxes are of great importance for training object detectors in the weakly supervised setting. 
In contrast to applying fully supervised segmentation models~\cite{shi2017weakly}, or training additional weakly supervised segmentation models~\cite{bearman2016s,kolesnikov2016seed}, we incorporate variants of the GrabCut method 
as segmentation cues to refine object bounding boxes obtained by MIL.
\item We present detailed evaluations on the PASCAL VOC and ILSVRC datasets. 
Experimental results demonstrate that our progressive representation adaptation algorithm performs favorably against the state-of-the-art methods. 
We also present comprehensive ablation studies to show the effect of each component, validating the non-trial algorithmic combinations and designs.
\end{compactenum}


\section{Related Work}

{\flushleft {\bf Multiple instance learning based methods.}}
Existing methods often cast WSL as a MIL problem~\cite{song2014learning, cinbis2017weakly, bilen2015weakly, bilen2014weakly, siva2012defence, siva2011weakly, siva2013looking, song2014weakly}.
MIL based methods alternate between selecting positive instances from positive images and learning object category classifiers.
However, the formulation results in a non-convex optimization problem. 
Due to non-convex objective functions, MIL based methods are sensitive to model initialization and prone to getting trapped into local extrema. 
Although significant efforts have been made to alleviate this problem via seeking better initialization models~\cite{song2014learning, song2014weakly, siva2013looking, siva2011weakly, siva2012defence} and optimization strategies~\cite{cinbis2017weakly, bilen2015weakly, bilen2014weakly}, the accuracy of selected object instances is still limited.
Our main observation is that existing MIL based methods attempt to train object detectors directly from the large and noisy collection of object candidates. This limits the quality of selected object proposals and the performance of object detectors.
In contrast, we propose to \emph{progressively} select good object candidates and transfer the classification network to a detection network. 
The proposed approach also applies MIL~\cite{song2014learning} to mine confident candidates.
However, unlike existing methods, we apply MIL on a cleaner collection of \emph{class-specific} object proposals instead of on a large, noisy, class-independent proposals. 
Class-specific object proposal mining helps alleviate the impact of background clutter and confusion from other categories. 
Furthermore, it can significantly reduce the computation cost of MIL training as the candidate set becomes smaller.

{\flushleft {\bf Neural networks for object localization.}}
Convolutional neural networks have recently achieved great success on various visual recognition tasks~\cite{krizhevsky2012imagenet, szegedy2014going, wei2014cnn, simonyan2014very, sermanet2013overfeat, girshick2014rich, girshick2015fast}. 
The key ingredient for the success is the end-to-end training in a \emph{fully supervised} fashion. 
For training, these CNN based methods~\cite{sermanet2013overfeat, girshick2014rich, girshick2015fast, ren2015faster} require a large number of instance-level annotations.
Moving beyond strong supervision, recent work focuses on applying off-the-shelf CNN features~\cite{song2014learning, song2014weakly, wang2014weakly, Bazzani:WACV16, bilen2015weakly, bilen2014weakly}, deriving object locations from feature maps of classification network~\cite{oquab2015object, singh2017hide, zhou2016learning, bency2016weakly}, object proposal mining~\cite{wang2014weakly, diba2016weakly, jie2017deep}, or training end-to-end weakly supervised detection network~\cite{bilen2016weakly, kantorov2016contextlocnet}.
In contrast to applying off-the-shelf CNN features, we first adapt a pre-trained network to perform multi-label image classification and use the mined object instances to train an object detector.
Recent work has shown that the network trained for image classification can also provide useful information for object localization~\cite{oquab2015object, singh2017hide, zhou2016learning, bency2016weakly}. 
Our classification adaptation step is conceptually similar to the method by Oquab~\etal~\cite{oquab2015object} in the formulation of multi-label classification. 
We use a different multi-label loss and extend the dimensions of the classification layer twice to output both of probabilities that an image contains objects from a certain category or not.
Furthermore, we focus on detecting the locations \emph{and} the spatial supports of objects while the method by Oquab~\etal~\cite{oquab2015object} only predicts approximate locations of objects. 
Our class-specific object proposal mining resembles the work by Bazzani~\etal~\cite{Bazzani:WACV16}. 
The main differences are three-fold. 
(1) We compute contrast scores for ranking proposals based on the region itself and its mask-out image.
(2) Instead of training a classifier over pre-trained CNN features, we fine-tune the parameters of all the layers for adapting the classification network into a detection network. 
(3) We combine activation scores derived from class-specific feature maps~\cite{zhou2016learning} for mining more accurate proposals.
Recent methods also use object proposal mining for WSL, including latent category learning~\cite{wang2014weakly} and dense subgraph discovery~\cite{jie2017deep}. 
Different from these methods, we propose to progressively mine high-quality object candidates for training object detectors.

{\flushleft {\bf Domain adaptation based methods.}}
Several recent approaches adopt domain adaptation and transfer learning techniques to 
facilitate learning object-level features or detectors~\cite{rochan2015weakly, hoffman2014lsda, hoffman2014detector, shi2012transfer, shi2016weakly, shi2017weakly, kumar2016track}. 
Prior knowledge can be extracted from the source domain and transferred to the target domain. 
Examples of source knowledge include the mapping relationship between the bounding box overlap and appearance similarity~\cite{shi2012transfer}, object size prior~\cite{shi2016weakly}, the appearance of similar objects~\cite{rochan2015weakly}, and tightness of bounding boxes of tracked objects in video~\cite{kumar2016track}. 
While domain adaptation and transfer learning techniques have shown promising results on WSL, these methods often require additional data with bounding box annotations to extract the useful prior knowledge for object localization. 
%
%
Our domain adaptation algorithm differs from these existing approaches in that we focus on object localization in a weakly supervised setting, i.e., we do not require any additional strongly annotated data or pre-trained detectors for similar object categories.

{\flushleft {\bf Progressive and self-paced learning.}} 
Our work is also related to several progressive and self-paced learning algorithms in other problem contexts. 
Examples include visual tracking~\cite{supancic2013self}, pose estimation~\cite{ferrari2008progressive}, image search~\cite{jiang2014easy}, and object discovery~\cite{lee2011learning}. 
Progressive methods can decompose complex problems into simpler ones. 
We note that progressive adaptation is of particular importance to the weakly supervised object localization problem.


\begin{figure}[t]
\centering
\includegraphics[width=\linewidth]{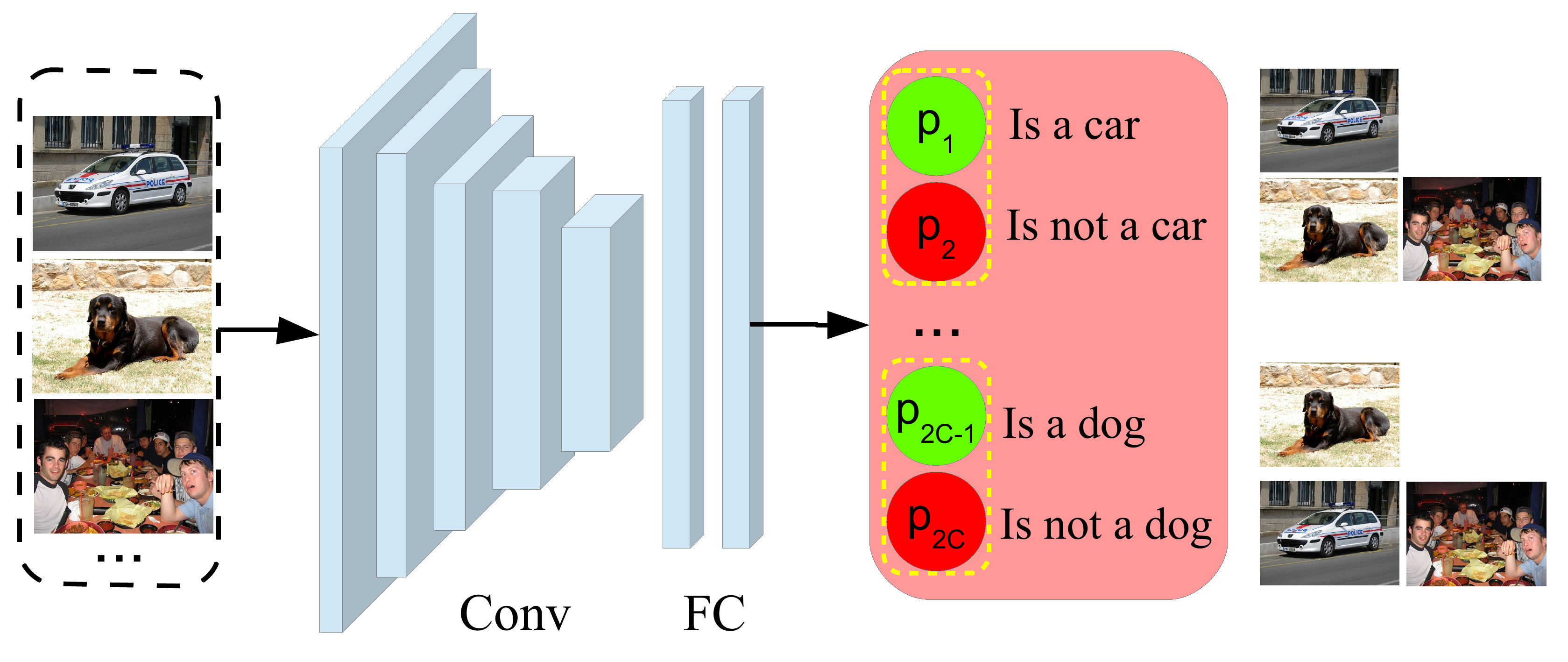}
\caption{\textbf{Classification adaptation}. 
%
We set the number of output nodes in the last fully-connected layer to $2C$, where $C$ is number of object categories. 
These $2C$ entries are grouped into $C$ pairs for indicating the \emph{presence} and \emph{absence} of each object category.
See Section~\ref{section:cls-adapt} for details.
}
\label{figure:cls-adapt}
\end{figure}

\section{Classification Adaptation}
\label{section:cls-adapt}

\begin{figure*}[!t]
\centering
\includegraphics[width=\linewidth]{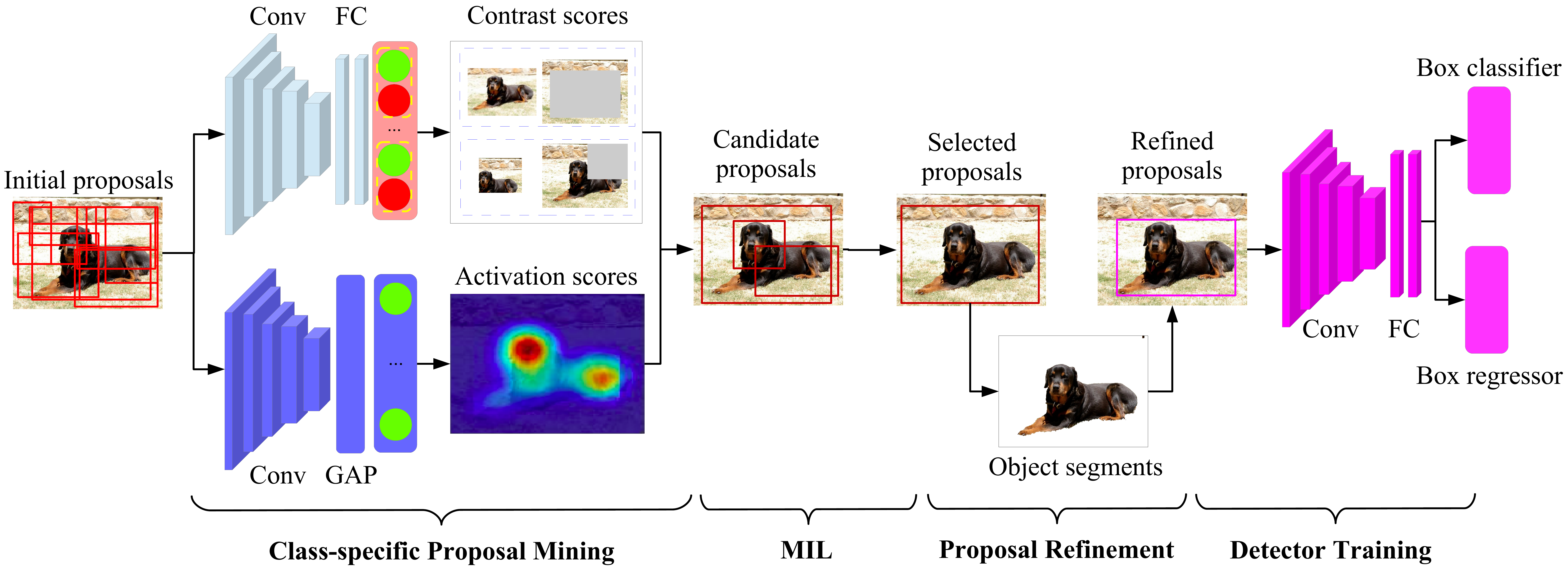}
\caption{\textbf{Detection adaptation}. 
We start with generating a collection of class-independent proposal using an off-the-shelf object proposal algorithm~\cite{zitnick2014edge}.
We leverage two scoring strategies to collect class-specific object proposals from this set of class-independent proposals (Section~\ref{section:mining}). 
The \emph{contrast scores} (\ref{equation:contrast}) measure the class prediction confidence drop between an object proposal and its mask-out image using adapted classification network.
The activation scores (\ref{equation:activation}) are computed based on class-specific feature maps~\cite{zhou2016learning}. 
After ranking the proposals using the two scores, we then apply multiple instance learning to select confident candidates for each class (Section~\ref{section:mil}).
We further refine the selected proposals using segmentation cues to obtain bounding boxes with tight spatial support (Section~\ref{section:refine}).
Finally, we use the refined object proposals to fine-tune all the layers (marked magenta), resulting in a network that is fully adapted for detection (Section~\ref{section:detector}).
}
\label{figure:det-adapt}
\end{figure*}

In this section, we introduce the classification adaptation step. 
This step aims to train the multi-label image classification network to increase the specificity of the adapted representation to the object categories of interest.
The original AlexNet~\cite{krizhevsky2012imagenet} is trained for multi-class classification with a softmax loss layer by assuming that only one single object exists per image. 
In our case, we replace the last classification layer with a multi-label loss layer.
Unlike the problem in ImageNet classification, we address a more general multi-label image classification problem where each image may contain multiple objects from more than one category.

Assuming that the object detection dataset has $C$ categories and a total of $N$ training images, we denote the weakly labeled training image set as $\mathcal{I} = \{(\bI^{(1)}, \by^{(1)}),\ldots,(\bI^{(N)}, \by^{(N)})\}$, where $\bI$ is the image data and $\by = [y_1,\ldots,y_c,\ldots,y_C]^\top \in \{0,1\}^C, c\in\{1,\ldots,C\}$ is the $C$-dimensional label vector of $\bI$. Each entry of $\by$ can be $1$ or $0$ indicating whether at least one specific object instance is present in the image. 
In the weakly object localization setting, one image may contain objects from different categories, i.e., more than one entry in $\by$ can be 1. 
In this case, conventional \emph{softmax} loss cannot be used for this multi-label classification problem. 
We thus introduce a multi-label loss to handle this problem.

First, we convert the original training label to a new label $\bt \in \{0,1\}^{2C}$, where
\begin{equation}
t_{2c-1} = \left \{
\begin{aligned}
1,\ y_c = 1\\
0,\ y_c = 0
\end{aligned}
\right.
\quad
\text{and}
\quad
t_{2c} = \left \{
\begin{aligned}
0,\ y_c = 1\\
1,\ y_c = 0
\end{aligned}
\right. .
\end{equation}
In other words, each odd entry of $\bt$ represents whether the image contains the corresponding object.
Similarly, each even entry represents whether the image \emph {does not} contain the corresponding object.

We now describe the proposed loss layer for multi-label classification. 
We denote the CNN as a function $\bp(\cdot)$ that maps an input image $\bI$ to a $2C$ dimensional output $\bp(\bI) \in \bbR^{2C}$. 
The odd entry $p_{2c-1}(\bI)$ represents the probability that the image contains at least one object instance of $c$-th category. 
Similarly, the even entry $p_{2c}(\bI)$ indicates the probability that the image does not contain objects of $c$-th category. 
We compute the probabilities using a sigmoid for each object class and thus we have $p_{2c-1}(\bI) + p_{2c}(\bI) = 1$. 

We can define negative logarithmic classification loss $L_c(\bI)$ of one image for category $c$ as,
\begin{equation}
L_c(\bI) = -(t_{2c-1}\log p_{2c-1}(\bI) + t_{2c}\log p_{2c}(\bI)).
\end{equation}
We obtain the final loss function $L$ by summing up all the training samples and losses for all the categories:
\begin{equation}
L =\sum_{i=1}^{N}\sum_{c = 1}^{C} L_c(\bI^{(i)}) = -\sum_{i = 1}^{N}\bt^{(i)} \cdot \log \bp(\bI^{(i)}).
\end{equation}
Here $\log(\cdot)$ is the element-wise logarithmic function. 

In the classification adaptation network, we substitute the conventional softmax loss layer with the multi-label loss layer and adjust the number of nodes in the last fully-connected layer to $2C$. 
We use mini-batch Stochastic Gradient Descent (SGD) for training the classification network. 
We initialize all the layers except the last layer using the pre-trained parameters on ImageNet~\cite{deng2009imagenet}.
For the modified classification layer, we randomly initialize the weights.
Further implementation details are described in Section~\ref{section:implementation}.


\section{Detection Adaptation}
\label{section:det-adapt}

\subsection{Class-specific proposal mining}
\label{section:mining}

The goal of detection adaptation is to transfer the multi-label image classifiers to object detectors. 
To train the object detectors, we first collect confident object proposals. 
We propose two strategies to collect \emph{class-specific} object proposals and apply multiple instance learning to mine confident candidates. 
The mining procedure offers two key benefits:
\begin{compactitem} 
\item Compared with class-independent object proposals, class-specific proposals remove substantial noise and potential confusion from similar objects.
This helps MIL avoid converging to an undesirable local minimum and reduce computational complexity.
\item More precise object proposals can be mined by MIL after class-specific proposal mining. 
These confident object proposals allow us to further fine-tune the network for object detection.
\end{compactitem}

{\flushleft {\bf Contrast scores.}}
The adapted classification network from Section~\ref{section:cls-adapt} predicts whether an input image contains a certain object category.
We use a mask-out strategy to collect object proposals for each class based on the adapted classification network.
The idea of masking out the input of CNN has been previously explored in~\cite{zeiler2014visualizing, Bazzani:WACV16, singh2017hide}. 
Intuitively, if the mask-out image by a region causes a significant drop in classification score for the $c$-th class, the region can be considered discriminative for the $c$-th class. 
Similar to~\cite{zeiler2014visualizing,Bazzani:WACV16}, we exploit the contrastive relationship between a selected region and its mask-out image.

Without loss of generality, we take mining object proposals for the $c$-th category as an example. 
First, for the image $\bI$, we apply the Edge Box algorithm~\cite{zitnick2014edge} to generate the initial collection of object proposals. 
The set of initial proposals is marked as $\mathbf{\hB}$. 
For an initial bounding box $\mathbf{\hb}$, we denote the region image as $\bI_\mathrm{in}(\mathbf{\hb})$ and its mask-out image as $\bI_\mathrm{out}(\mathbf{\hb})$. 
The mask-out image is generated by replacing the pixel values within $\mathbf{\hb}$ with the fixed mean pixel values pre-computed on the ILSVRC 2012 dataset. 
We feed each region image $\bI_\mathrm{in}(\mathbf{\hb})$ and the corresponding mask-out image $\bI_\mathrm{out}(\mathbf{\hb})$ to the adapted classification network. 
We can then compute the contrast score for each bounding box $\mathbf{\hb}$ of image $\bI$ as 
\begin{equation}
\mathrm{Contrast}_c(\mathbf{\hb}) = p_{2c-1}(\bI_\mathrm{in}(\mathbf{\hb})) - p_{2c-1}(\bI_\mathrm{out}(\mathbf{\hb})).
\label{equation:contrast}
\end{equation}
Here, if the value of $\mathrm{Contrast}_c(\mathbf{\hb})$ is large, it indicates that the region $\mathbf{\hb}$ is likely an object of the $c$-th category.
Note that our mask-out strategy differs from~\cite{Bazzani:WACV16}, which computes the score difference between the whole image and mask-out image. 

{\flushleft {\bf Activation scores.}} Recent work has shown that the convolutional layers can be used to localize objects or discriminative regions in the classification network~\cite{zhou2016learning, singh2017hide}. 
We also exploit class activation maps~\cite{zhou2016learning} to help select class-specific proposals for each class.

For a given image $\bI$, we denote $a_k(x,y)$ as its activation of the $k$-th feature map at the position of $(x,y)$ in the last convolutional layer.
The confidence for class $c$ can be computed as
\begin{equation}
\footnotesize
\sum_{k}w_k\sum_{x,y}a_k(x,y) = \sum_{x,y}\sum_{k}w_k \cdot a_k(x,y) = \sum_{x,y}m_c(x,y),
\label{equation:cam}
\end{equation}
where $w_k^c$ is the weight to class $c$ for the $k$-th feature map and $m_c(x,y) = \sum_{k}w_k^c \cdot a_k(x,y)$ is the activation map for class $c$. Thus, $m_c(x,y)$ indicates the importance of the network activation at $(x, y)$ for recognizing the $c$-th category.

However, due to the lack of object-level supervision, the class activation map $m_c$ alone is not sufficient to obtain accurate object bounding boxes.
Instead of directly deriving candidate proposals by thresholding the class-specific heat map, 
we compute activation scores for each of class-independent object proposals based on $m_c$ and then select top ones for class $c$. 
To this end, we first compute the integral image of the class activation map, $H_c(x,y) = \sum_{x'<x,y'<y}m_c(x',y')$. 
The response of a bounding box $\mathbf{\hb}$ to locate objects from the $c$-th category can then be efficiently computed using
\begin{equation}
\footnotesize
r_c(\mathbf{\hb}) = H_c(x_1,y_1)+H_c(x_2,y_2)-H_c(x_1,y_2)-H_c(x_2,y_1),
\label{equation:response}
\end{equation}
where $(x_1,y_1)$ denotes the upper-left coordinate of $\mathbf{\hb}$ and $(x_2,y_2)$ denotes its bottom-right coordinate.

The response $r_c$, however, often underestimates the size of objects.
We thus incorporate an object size prior to obtain the activation score:
\begin{equation}
\mathrm{Activation}_c(\mathbf{\hb}) = \frac{r_c(\mathbf{\hb})}{wh}+\alpha \cdot \frac{r_c(\mathbf{\hb})}{H_c(W,H)},
\label{equation:activation}
\end{equation}
where $w/h$ denotes the width/length of the bounding box $\mathbf{\hb}$ and $W/H$ denotes the width/length of the whole image $\bI$.
We select $\alpha$ with highest recall on the validation set using parameter sweeping ($\alpha=5$ in our experiments).
Here, if the value of $\mathrm{Activation}_c(\mathbf{\hb})$ is large, it indicates that the region $\mathbf{\hb}$ is likely an object of the $c$-th category. 

\begin{figure}[t]
\centering
\includegraphics[width=\linewidth]{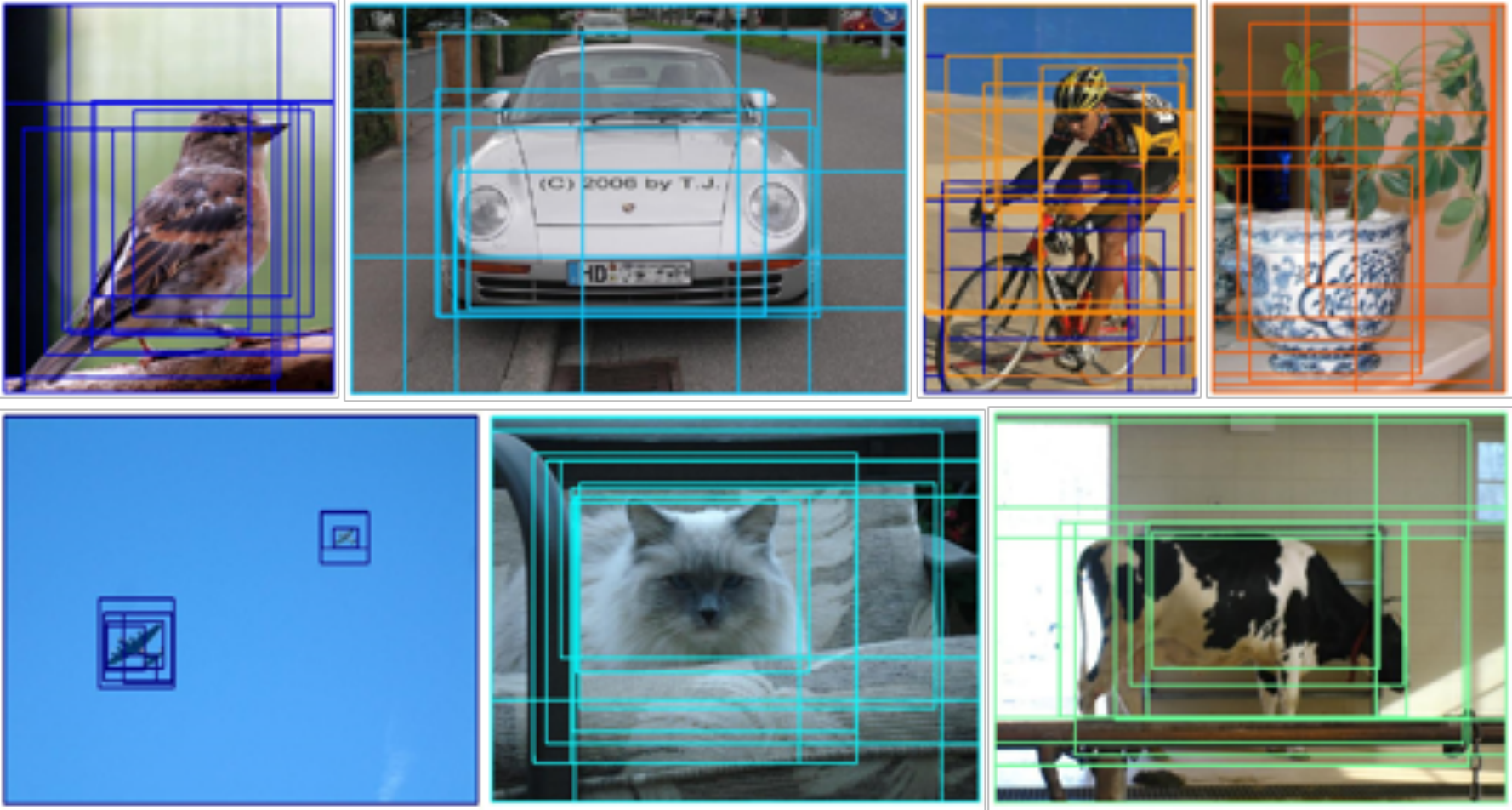} 
\caption{\textbf{Examples of class-specific object proposals}. We show top 10 proposals for each category (different colors indicate mined proposals for different categories).}
\label{figure:top10}
\end{figure}

{\flushleft {\bf Fusing contrast and activation scores.}}
The contrast and activation scores for an object proposal are complementary because they are derived from different levels of representations. 
Contrast scores are computed based on the classification layer of the adapted classification network.
Activation scores are computed based on the last convolutional layers.
We normalize both scores to the range of $[0, 1]$ and fuse these scores using a linear combination for ranking the class-specific object proposals.
We determine the weight with the highest recall on the validation set using parameter sweeping. 
We set the ratio of contrast score to activation score to $10:1$ in our experiments. 
According to the fused scores, we then select top $M$ proposals for each class. 
We mark the top-ranked class-specific proposals as $\bB_c$.
In Figure~\ref{figure:top10}, we show some examples of the mined class-specific proposals. 
The proposed strategies help mine proposals that are concentrated on and around the objects.

\subsection{Multiple instance learning}
\label{section:mil}

As the purpose of class-specific object mining aims to maintain high recall, the top-ranked proposals may still be coarse and contain many false positives. 
We thus apply MIL to further select confident candidates from the class-specific proposals. 
In MIL, the label of object candidate is set as a latent variable. 
During the training, the label is iteratively updated. 
%
For the candidate set $\bB_c$, we set up latent variable $\bz^{k}_c \in \{0,1\}^M, k, c \in {1,\ldots,C}$, where each entry in $\bz^{k}_c$ represents whether the corresponding proposal is an object of the $k$-th category.
We make two assumptions for solving $\bz^{k=c}_c$:
\begin{compactitem}
\item For image $\bI$ with $y_c = 1$, at least one proposal in $\bB_c$ belongs to the $c$-th category, i.e., $\mathbf{1}^\top \cdot \bz^{k=c}_c \geq 1$ where $\mathbf{1}$ is an $M$-dimensional all-one vector.
\item For image $\bI'$ with $y_c = 0$, none of proposals in $\bB'_{c'\neq c}$ belongs to the $c$-th category, i.e., $\mathbf{1}^\top \cdot \bz^{k = c}_{c'\neq c} = 0$.
\end{compactitem}
Under these two assumptions, we can treat each image with $y_c=1$ as a positive bag and treat each image with $y_c=0$ as a negative bag. 
We then cast the task of solving $\bz^{k=c}_c$ as an MIL problem. 
Note that multiple positive instances can be collected according to the scores of the MIL classifier for each class.

We use the smoothed hinge loss function in~\cite{song2014learning}. 
Note that the initialization step in \cite{song2014learning} is carried out via a sub-modular clustering method from the initial object proposals. 
The noisy collection of proposals substantially limits the performance of clustering process. 
In addition, the initialization step is time-consuming as the similarity measures among all the proposals in all the images need to be computed. 
With class-specific proposal mining, we can filter the object proposal collection and significantly reduce the training time of MIL. 
The run-time cost of our initialization step is $M/|\mathbf{\hb}|$ of that by~\cite{song2014learning} for each class.

\subsection{Bounding box refinement}
\label{section:refine} 

\begin{figure}[t]
\centering
\includegraphics[width=\linewidth]{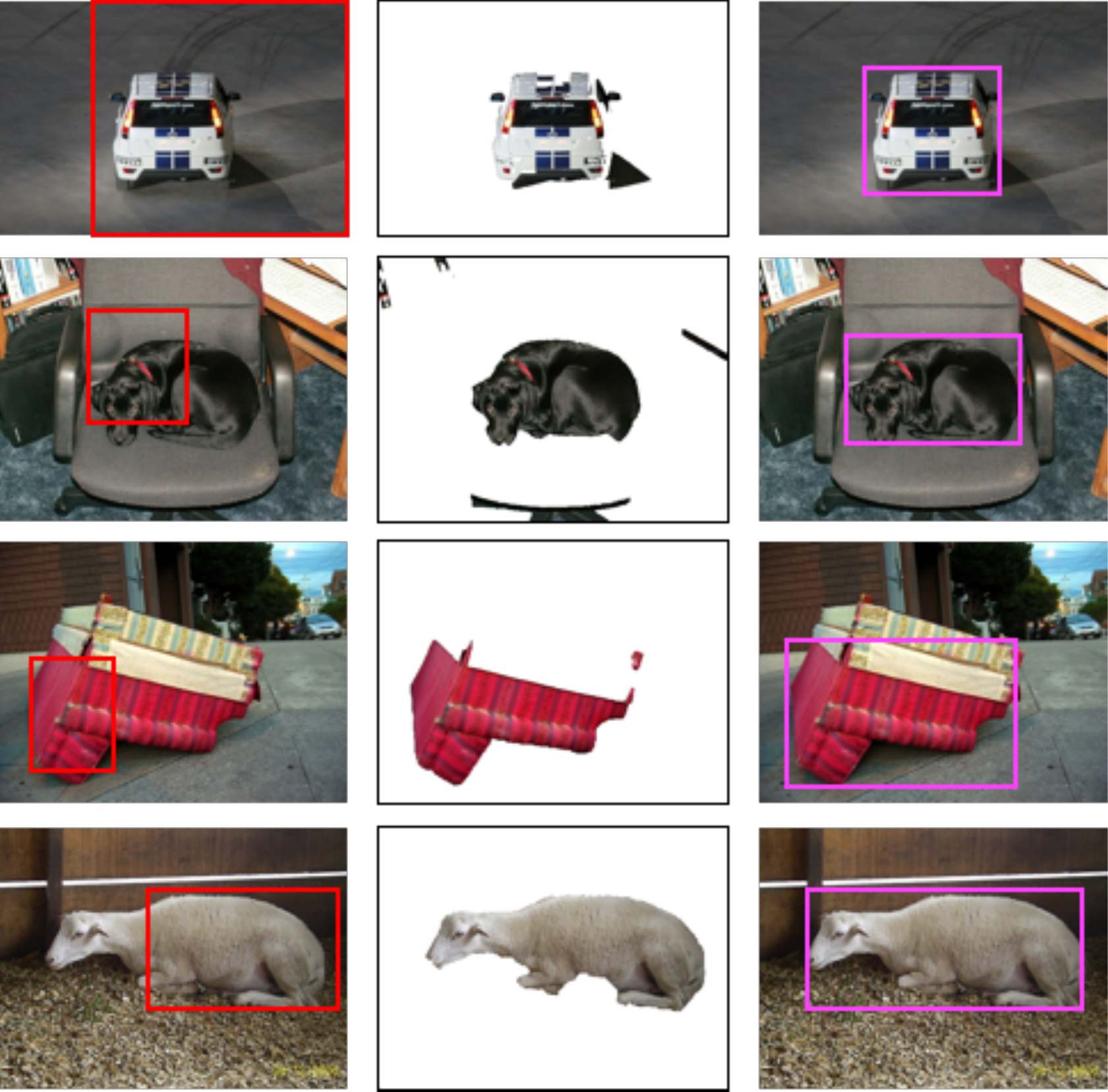}
\caption{\textbf{Examples of the refined bounding boxes}. The first column shows the selected proposals by MIL. The second column shows the generated segments using segmentation cues~\cite{tang2015secrets}. The last column shows the obtained tight bounding boxes. Using segmentation cues helps obtain improved localization accuracy.}
\label{figure:seg}
\end{figure}

For the task of object detection, the quality of bounding boxes is of critical importance. 
Before training an object detector, we use segmentation cues to refine the object bounding boxes mined by MIL.
Specifically, we apply the kernel GrabCut method~\cite{tang2015secrets} to segment objects given an initial bounding box. 
We then take the tightest bounding boxes enclosing the foreground segments as our refined bounding boxes. 
Figure~\ref{figure:seg} shows several examples of refined bounding boxes.

\subsection{Object detector learning}
\label{section:detector} 

In this step, we aim at adapting the network from multi-label image classification for object detection. 
We jointly train the detection network with $C$ object classes and a background class.
Similar to~\cite{girshick2015fast}, we replace the $2C$-dimensional classification layer (for image-level classification) with a randomly initialized ($C$+1)-dimensional classification layer (for instance-level classification with $C$ object classes and background). 
The smooth $L_1$ loss is used for bounding box regression. 
We take the refined object proposals as positive samples for each object category. 
Next, we collect background samples from object proposals that have a maximum IoU $\in [0.1,0.5)$ overlap with the mined object proposals by MIL. 
For data augmentation, we treat all the proposals that have IoU $\geq0.5$ overlap with a mined object as positive samples.

Given a test image, we first generate object proposals using the Edge Box method~\cite{zitnick2014edge} and use the adapted detection network to score each proposal. 
We then rank all the proposals and use non-maximum suppression to obtain the final detection.


\section{Experiments}

In this section, we first describe implementation details (Section~\ref{section:implementation}), datasets, and metrics for evaluation (Section~\ref{section:metrics}).
We then present quantitative comparisons with the state-of-the-art weakly supervised object localization algorithms (Section~\ref{section:comparisons}).
To better understand the contribution of each component in the proposed approach, we conduct an ablation study (Section~\ref{section:ablation}).
Finally, we examine the sensitivity of important hyper-parameters used in our method (Section~\ref{section:parameter}) and present detector error analysis (Section~\ref{section:error}).

\subsection{Implementation details}
\label{section:implementation}

For multi-label image classification training, we use the AlexNet~\cite{krizhevsky2012imagenet} pretrained on the ImageNet as our base CNN model. 
We train the network with SGD at a learning rate of 0.001 for 10,000 mini-batch iterations. 
We set the size of mini-batch to 500.
For class-specific proposal mining, we use the Edge Box method~\cite{zitnick2014edge} to generate 2,000 initial object proposals and select top 
50 proposals as the input for the multiple instance learning algorithm.
For object detector training, we take either the AlexNet~\cite{krizhevsky2012imagenet} or VGGNet~\cite{simonyan2014very} as our base models in the Fast-RCNN framework~\cite{girshick2015fast}.
Similar to the Fast-RCNN~\cite{girshick2015fast}, we set the maximum number of iterations to 40,000.

We implement the proposed algorithm using the Caffe toolbox~\cite{jia2014caffe}. 
For the PASCAL VOC 2007 \emph{trainval} set, it takes about 10 hours to fine-tune the AlexNet for classification with a Tesla K40 GPU.
The detection adaptation task takes about 1 hour.
With the mined class-specific proposals, it takes about 3 hours to select confident object instances for each class on PC with a 4.0 GHz Intel i7 CPU and 16 GB memory.
The source code will be made available to the public. 

\subsection{Datasets and evaluation metrics}
\label{section:metrics}

{\flushleft {\bf Datasets.}}
We extensively evaluate the proposed method on the PASCAL VOC 2007, 2010, 2012~\cite{everingham2010pascal, everingham2014pascal} and ILSVRC 2013 detection 
datasets~\cite{deng2009imagenet,ILSVRC15}. 
For the PSCAL VOC datasets, we use both \emph{train} and \emph{val} splits as the training set and use the \emph{test} split as our test set.
For the ILSVRC 2013 detection dataset, we follow the RCNN~\cite{girshick2014rich} to split the \emph{val} data into \emph{val$_1$} and \emph{val$_2$}. 
We use the training data in \emph{val$_1$} for training object detectors and \emph{val$_2$} for validating the localization performance. 
Note that we do not use any instance-level annotations (i.e., object bounding boxes) for training in all the datasets.

{\flushleft {\bf Evaluation metrics.}}
We use two metrics to evaluate the localization performance. 
First, we compute the fraction of positive training images in which we obtain correct localization (CorLoc)~\cite{deselaers2012weakly}.
Second, we measure the performance of object detectors using average precision in the test set. 
For both metrics, we consider that a bounding box is correct if it the intersection-over-union (IoU) ratio with a ground-truth object instance annotation is at least 50\%.

\subsection{Comparisons to the state-of-the-art}
\label{section:comparisons}

\begin{table*}[!t]
\caption{\textbf{Quantitative comparisons in terms of correct localization on the PASCAL VOC 2007 \emph{trainval} set}.}
\label{table:corloc}
\LARGE
\centering
\resizebox{\linewidth}{!}{
\begin{tabular}{ccccc|cccccccccccccccccccc|c}
\toprule
\multicolumn{5}{l|}{Methods} & aero & bike & bird & boat & bottle & bus & car & cat & chair & cow & table & dog & horse & mbike & person & plant & sheep & sofa & train & tv & Avg. \\
\midrule
\multicolumn{5}{l|}{Siva \etal \cite{siva2012defence}} & 45.8 & 21.8 & 30.9 & 20.4 & 5.3 & 37.6 & 40.8 & 51.6 & 7.0 & 29.8 & 27.5 & 41.3 & 41.8 & 47.3 & 24.1 & 12.2 & 28.1 & 32.8 & 48.7 & 9.4 & 30.2 \\
\multicolumn{5}{l|}{Shi \etal \cite{shi2013bayesian}} & 67.3 & 54.4 & 34.3 & 17.8 & 1.3 & 46.6 & 60.7 & 68.9 & 2.5 & 32.4 & 16.2 & 58.9 & 51.5 & 64.6 & 18.2 & 3.1 & 20.9 & 34.7 & 63.4 & 5.9 & 36.2 \\
\multicolumn{5}{l|}{Bilen \etal \cite{bilen2015weakly}} & 66.4 & 59.3 & 42.7 & 20.4 & 21.3 & 63.4 & 74.3 & 59.6 & 21.1 & 58.2 & 14.0 & 38.5 & 49.5 & 60.0 & 19.8 & 39.2 & 41.7 & 30.1 & 50.2 & 44.1 & 43.7 \\
\multicolumn{5}{l|}{Wang \etal \cite{wang2014weakly}} & 80.1 & 63.9 & 51.5 & 14.9 & 21.0 & 55.7 & 74.2 & 43.5 & 26.2 & 53.4 & 16.3 & 56.7 & 58.3 & 69.5 & 14.1 & 38.3 & 58.8 & 47.2 & 49.1 & 60.9 & 48.5 \\
\multicolumn{5}{l|}{Cinbis \etal \cite{cinbis2017weakly}} & 65.3 & 55.0 & 52.4 & 48.3 & 18.2 & 66.4 & 77.8 & 35.6 & 26.5 & 67.0 & 46.9 & 48.4 & 70.5 & 69.1 & 35.2 & 35.2 & 69.6 & 43.4 & 64.6 & 43.7 & 52.0 \\
\multicolumn{5}{l|}{Kantorov \etal \cite{kantorov2016contextlocnet}} & 83.1 & 68.8 & 54.7 & 23.4 & 18.3 & 73.6 & 74.1 & 54.1 & 8.6 & 65.1 & 47.1 & 59.5 & 67.0 & 83.5 & 35.3 & 39.9 & 67.0 & 49.7 & 63.5 & 65.2 & 55.1 \\
\multicolumn{5}{l|}{Jie \etal \cite{jie2017deep}} & 72.7 & 55.3 & 53.0 & 27.8 & 35.2 & 68.6 & 81.9 & 60.7 & 11.6 & 71.6 & 29.7 & 54.3 & 64.3 & 88.2 & 22.2 & 53.7 & 72.2 & 52.6 & 68.9 & 75.5 & 56.1 \\
\multicolumn{5}{l|}{Bilen \etal \cite{bilen2016weakly}} & 65.1 & 58.8 & 58.5 & 33.1 & 39.8 & 68.3 & 60.2 & 59.6 & 34.8 & 64.5 & 30.5 & 43.0 & 56.8 & 82.4 & 25.5 & 41.6 & 61.5 & 55.9 & 65.9 & 63.7 & 53.5 \\
\multicolumn{5}{l|}{Diba \etal \cite{diba2016weakly}} & 83.9 & 72.8 & 64.5 & 44.1 & 40.1 & 65.7 & 82.5 & 58.9 & 33.7 & 72.5 & 25.6 & 53.7 & 67.4 & 77.4 & 26.8 & 49.1 & 68.1 & 27.9 & 64.5 & 55.7 & 56.7 \\
\midrule
CS & AS & MIL & Seg & FT & & & & & & & & & & & & & & & & & & & & \\
$\surd$ & & & & & 50.4 & 30 & 34.6 & 18.2 & 6.2 & 39.3 & 42.2 & 57.3 & 10.8 & 29.8 & 20.5 & 41.8 & 43.2 & 51.8 & 24.7 & 20.8 & 29.2 & 26.6 & 45.6 & 12.5 & 31.8 \\
$\surd$ & & $\surd$ & & & 64.3 & 54.3 & 42.7 & 22.7 & 34.4 & 58.1 & 74.3 & 36.2 & 24.3 & 50.4 & 11.0 & 29.2 & 50.5 & 66.1 & 11.3 & 42.9 & 39.6 & 18.3 & 54.0 & 39.8 & 41.2 \\
$\surd$ & & $\surd$ & & \bf{S} & 77.3 & 62.6 & 53.3 & 41.4 & 28.7 & 58.6 & 76.2 & 61.1 & 24.5 & 59.6 & 18.0 & 49.9 & 56.8 & 71.4 & 20.9 & 44.5 & 59.4 & 22.3 & 60.9 & 48.8 & 49.8 \\
$\surd$ & & $\surd$ & & \bf{L} & 78.2 & 67.1 & 61.8 & 38.1 & 36.1 & 61.8 & 78.8 & 55.2 & 28.5 & 68.8 & 18.5 & 49.2 & 64.1 & 73.5 & 21.4 & 47.4 & 64.6 & 22.3 & 60.9 & 52.3 & 52.4 \\
$\surd$ & $\surd$ & $\surd$ & & \bf{L} & 82.4 & 75.4 & 62.2 & 38.5 & 34.6 & 64.2 & 83.5 & 58.3 & 28.0 & 72.0 & 18.0 & 53.5 & 68.0 & 75.3 & 21.4 & 48.2 & 66.2 & 22.8 & 62.9 & 54.6 & 54.5 \\
$\surd$ & $\surd$ & $\surd$ & $\surd$ & \bf{L} & 84.0 & 77.0 & 64.2 & 41.4 & 34.0 & 69.9 & 87.1 & 67.1 & 36.6 & 78.0 & 25.0 & 55.3 & 71.1 & 84.5 & 21.2 & 62.0 & 69.8 & 24.5 & 69.7 & 57.0 & 59.0 \\
\bottomrule
\end{tabular}
}
\end{table*}

\begin{table*}[!t]
\caption{\textbf{Quantitative comparisons in terms of average precision on the PASCAL VOC 2007 \emph{test} set}.}
\label{table:ap}
\LARGE
\centering
\resizebox{\linewidth}{!}{
\begin{tabular}{ccccc|cccccccccccccccccccc|c}
\toprule
\multicolumn{5}{l|}{Methods} & aero & bike & bird & boat & bottle & bus & car & cat & chair & cow & table & dog & horse & mbike & person & plant & sheep & sofa & train & tv & mAP \\
\midrule
\multicolumn{5}{l|}{Song \etal \cite{song2014learning}} & 27.6 & 41.9 & 19.7 & 9.1 & 10.4 & 35.8 & 39.1 & 33.6 & 0.6 & 20.9 & 10.0 & 27.7 & 29.4 & 39.2 & 9.1 & 19.3 & 20.5 & 17.1 & 35.6 & 7.1 & 22.7 \\
\multicolumn{5}{l|}{Song \etal \cite{song2014weakly}} & 36.3 & 47.6 & 23.3 & 12.3 & 11.1 & 36.0 & 46.6 & 25.4 & 0.7 & 23.5 & 12.5 & 23.5 & 27.9 & 40.9 & 14.8 & 19.2 & 24.2 & 17.1 & 37.7 & 11.6 & 24.6 \\
\multicolumn{5}{l|}{Bilen \etal \cite{bilen2014weakly}} & 42.2 & 43.9 & 23.1 & 9.2 & 12.5 & 44.9 & 45.1 & 24.9 & 8.3 & 24.0 & 13.9 & 18.6 & 31.6 & 43.6 & 7.6 & 20.9 & 26.6 & 20.6 & 35.9 & 29.6 & 26.4 \\
\multicolumn{5}{l|}{Bilen \etal \cite{bilen2015weakly}} & 46.2 & 46.9 & 24.1 & 16.4 & 12.2 & 42.2 & 47.1 & 35.2 & 7.8 & 28.3 & 12.7 & 21.5 & 30.1 & 42.4 & 7.8 & 20.0 & 26.8 & 20.8 & 35.8 & 29.6 & 27.7 \\
\multicolumn{5}{l|}{Cinbis \etal \cite{cinbis2017weakly}} & 39.3 & 43.0 & 28.8 & 20.4 & 8.0 & 45.5 & 47.9 & 22.1 & 8.4 & 33.5 & 23.6 & 29.2 & 38.5 & 47.9 & 20.3 & 20.0 & 35.8 & 30.8 & 41.0 & 20.1 & 30.2 \\
\multicolumn{5}{l|}{Wang \etal \cite{wang2014weakly}} & 48.9 & 42.3 & 26.1 & 11.3 & 11.9 & 41.3 & 40.9 & 34.7 & 10.8 & 34.7 & 18.8 & 34.4 & 35.4 & 52.7 & 19.1 & 17.4 & 35.9 & 33.3 & 34.8 & 46.5 & 31.6 \\
\multicolumn{5}{l|}{Bilen \etal \cite{bilen2016weakly}} & 39.4 & 50.1 & 31.5 & 16.3 & 12.6 & 64.5 & 42.8 & 42.6 & 10.1 & 35.7 & 24.9 & 38.2 & 34.4 & 55.6 & 9.4 & 14.7 & 30.2 & 40.7 & 54.7 & 46.9 & 34.8 \\
\multicolumn{5}{l|}{Kantorov \etal \cite{kantorov2016contextlocnet}} & 57.1 & 52.0 & 31.5 & 7.6 & 11.5 & 55.0 & 53.1 & 34.1 & 1.7 & 33.1 & 49.2 & 42.0 & 47.3 & 56.6 & 15.3 & 12.8 & 24.8 & 48.9 & 44.4 & 47.8 & 36.3 \\
\multicolumn{5}{l|}{Jie \etal \cite{jie2017deep}} & 52.2 & 47.1 & 35.0 & 26.7 & 15.4 & 61.3 & 66.0 & 54.3 & 3.0 & 53.6 & 24.7 & 43.6 & 48.4 & 65.8 & 6.6 & 18.8 & 51.9 & 43.6 & 53.6 & 62.4 & 41.7 \\
\multicolumn{5}{l|}{Diba \etal \cite{diba2016weakly}} & 49.5 & 60.6 & 38.6 & 29.2 & 16.2 & 70.8 & 56.9 & 42.5 & 10.9 & 44.1 & 29.9 & 42.2 & 47.9 & 64.1 & 13.8 & 23.5 & 45.9 & 54.1 & 60.8 & 54.5 & 42.8 \\
\midrule
CS & AS & MIL & Seg & FT & & & & & & & & & & & & & & & & & & & & \\
$\surd$ & & $\surd$ & & & 37.2 & 35.7 & 25.8 & 13.8 & 12.7 & 36.2 & 42.4 & 22.3 & 14.3 & 24.2 & 9.4 & 13.1 & 27.9 & 38.9 & 3.7 & 18.7 & 20.1 & 16.3 & 36.1 & 18.4 & 23.4 \\
$\surd$ & & & & \bf{S} & 30.4 & 22.4 & 15.0 & 3.5 & 2.8 & 26.6 & 27.3 & 46.8 & 0.8 & 10.8 & 13.1 & 34.7 & 35.8 & 38.7 & 12.6 & 8.4 & 8.8 & 12.8 & 33.6 & 4.6 & 19.5 \\
& & $\surd$ & & \bf{S} & 17.5 & 50.2 & 22.5 & 4.0 & 9.9 & 38.8 & 48.7 & 39.3 & 0.3 & 22.1 & 10.1 & 19.8 & 22.4 & 49.9 & 3.4 & 15.5 & 32.1 & 10.8 & 40.0 & 1.9 & 23.0 \\
$\surd$ & & $\surd$ & & \bf{S} & 49.7 & 33.6 & 30.8 & 19.9 & 13 & 40.5 & 54.3 & 37.4 & 14.8 & 39.8 & 9.4 & 28.8 & 38.1 & 49.8 & 14.5 & 24.0 & 27.1 & 12.1 & 42.3 & 39.7 & 31.0 \\
$\surd$ & $\surd$ & $\surd$ & & \bf{S} & 46.1 & 32.8 & 30.6 & 16.1 & 12.3 & 42.4 & 51.7 & 46.3 & 12.6 & 29.0 & 20.3 & 37.3 & 36.3 & 47.3 & 14.4 & 24.2 & 30.0 & 19.3 & 33.3 & 52.0 & 31.7 \\
$\surd$ & $\surd$ & $\surd$ & $\surd$ & \bf{S} & 50.1 & 28.7 & 35.6 & 12.7 & 14.2 & 45.3 & 65.3 & 59.5 & 14.7 & 26.7 & 11.0 & 47.6 & 29.3 & 50.5 & 13.5 & 25.5 & 21.8 & 17.6 & 33.8 & 57.9 & 33.1 \\
$\surd$ & & & & {L} & 30.4 & 25.3 & 11.1 & 6.3 & 1.5 & 31.3 & 29.4 & 49.1 & 1.0 & 10.6 & 12.6 & 42.0 & 38.7 & 36.7 & 12.8 & 10.8 & 10.3 & 10.3 & 34.1 & 5.0 & 20.5 \\
& & $\surd$ & & {L} & 25.6 & 58.5 & 25.3 & 1.8 & 11.7 & 43.5 & 53.4 & 35.7 & 0.2 & 32.3 & 10.7 & 19.3 & 32.8 & 56.5 & 1.8 & 15.6 & 37.3 & 16.0 & 43.6 & 2.9 & 26.2 \\
$\surd$ & & $\surd$ & & \bf{L} & 54.5 & 47.4 & 41.3 & 20.8 & 17.7 & 51.9 & 63.5 & 46.1 & 21.8 & 57.1 & 22.1 & 34.4 & 50.5 & 61.8 & 16.2 & 29.9 & 40.7 & 15.9 & 55.3 & 40.2 & 39.5 \\
$\surd$ & $\surd$ & $\surd$ & & \bf{L} & 58.3 & 65.7 & 49.0 & 20.8 & 13.0 & 61.5 & 66.7 & 52.1 & 21.0 & 61.3 & 20.0 & 31.5 & 51.0 & 65.2 & 3.2 & 25.3 & 48.3 & 17.3 & 59.3 & 34.0 & 41.2 \\
$\surd$ & $\surd$ & $\surd$ & $\surd$ & \bf{L} & 60.7 & 66.2 & 49.1 & 19.8 & 15.5 & 60.9 & 67.8 & 54.5 & 20.5 & 62.4 & 26.4 & 27.7 & 56.3 & 65.5 & 3.8 & 26.4 & 50.8 & 15.9 & 62.3 & 37.9 & 42.5 \\
\bottomrule
\end{tabular}
}
\end{table*}

\begin{table*}[!t]
\caption{\textbf{Quantitative comparisons in terms of average precision on the PASCAL VOC 2010 \emph{test} set}.}
\label{table:ap-2010}
\LARGE
\centering
\resizebox{\linewidth}{!}{
\begin{tabular}{ccccc|cccccccccccccccccccc|c}
\toprule
\multicolumn{5}{l|}{Methods} & aero & bike & bird & boat & bottle & bus & car & cat & chair & cow & table & dog & horse & mbike & person & plant & sheep & sofa & train & tv & mAP \\
\midrule
\multicolumn{5}{l|}{Cinbis \etal \cite{cinbis2017weakly}} & 44.6 & 42.3 & 25.5 & 14.1 & 11.0 & 44.1 & 36.3 & 23.2 & 12.2 & 26.1 & 14.0 & 29.2 & 36.0 & 54.3 & 20.7 & 12.4 & 26.5 & 20.3 & 31.2 & 23.7 & 27.4 \\
\multicolumn{5}{l|}{Bilen \etal \cite{bilen2016weakly}} & 57.4 & 51.8 & 41.2 & 16.4 & 22.8 & 57.3 & 41.8 & 34.8 & 13.1 & 37.6 & 10.8 & 37.0 & 45.2 & 64.9 & 14.1 & 22.3 & 33.8 & 27.6 & 49.1 & 44.8 & 36.2 \\
\multicolumn{5}{l|}{Diba \etal \cite{diba2016weakly}} & - & - & - & - & - & - & - & - & - & - & - & - & - & - & - & - & - & - & - & - & 39.5 \\
\midrule
CS & AS & MIL & Seg & FT & & & & & & & & & & & & & & & & & & & & \\
$\surd$ & & $\surd$ & & \bf{S} & 41.6 & 32.0 & 21.5 & 6.9 & 9.3 & 47.1 & 32.6 & 35.4 & 8.1 & 20.1 & 3.8 & 22.0 & 26.8 & 45.5 & 8.9 & 11.8 & 24.4 & 7.7 & 29.9 & 30.1 & 23.3 \\
$\surd$ & & $\surd$ & & \bf{L} & 54.2 & 49.1 & 38.5 & 11.4 & 18.7 & 56.0 & 44.6 & 43.3 & 14.5 & 41.3 & 7.9 & 35.3 & 49.9 & 63.2 & 10.4 & 17.4 & 38.3 & 15.1 & 45.9 & 37.8 & 34.6 \\
$\surd$ & $\surd$ & $\surd$ & $\surd$ & \bf{L} & 63.5 & 55.7 & 45.0 & 15.9 & 14.6 & 57.6 & 52.7 & 50.8 & 14.0 & 44.3 & 3.9 & 30.6 & 55.9 & 68.2 & 4.0 & 22.7 & 40.7 & 5.8 & 52.3 & 28.2 & 36.3 \\
\bottomrule
\end{tabular}
}
\end{table*}

\begin{table*}[!t]
\caption{\textbf{Quantitative comparisons in terms of average precision on the PASCAL VOC 2012 \emph{test} set}.}
\label{table:ap-2012}
\LARGE
\centering
\resizebox{\linewidth}{!}{
\begin{tabular}{ccccc|cccccccccccccccccccc|c}
\toprule
\multicolumn{5}{l|}{Methods} & aero & bike & bird & boat & bottle & bus & car & cat & chair & cow & table & dog & horse & mbike & person & plant & sheep & sofa & train & tv & mAP \\
\midrule
\multicolumn{5}{l|}{Kantorov \etal \cite{kantorov2016contextlocnet}} & 64.0 & 54.9 & 36.4 & 8.1 & 12.6 & 53.1 & 40.5 & 28.4 & 6.6 & 35.3 & 34.4 & 49.1 & 42.6 & 62.4 & 19.8 & 15.2 & 27.0 & 33.1 & 33.0 & 50.0 & 35.3 \\
\multicolumn{5}{l|}{Diba \etal \cite{diba2016weakly}} & - & - & - & - & - & - & - & - & - & - & - & - & - & - & - & - & - & - & - & - & 37.9 \\
\multicolumn{5}{l|}{Jie \etal \cite{jie2017deep}} & 60.8 & 54.2 & 34.1 & 14.9 & 13.1 & 54.3 & 53.4 & 58.6 & 3.7 & 53.1 & 8.3 & 43.4 & 49.8 & 69.2 & 4.1 & 17.5 & 43.8 & 25.6 & 55.0 & 50.1 & 38.3 \\
\midrule
CS & AS & MIL & Seg & FT & & & & & & & & & & & & & & & & & & & & \\
$\surd$ & & $\surd$ & & \bf{S} & 34.2 & 26.0 & 18.6 & 6.8 & 7.2 & 44.2 & 28.2 & 32.2 & 7.9 & 19.5 & 6.1 & 23.7 & 30.7 & 46.6 & 11.7 & 13.4 & 18.2 & 3.9 & 28.5 & 25.8 & 21.7 \\
$\surd$ & & $\surd$ & & \bf{L} & 50.7 & 43.0 & 31.0 & 12.1 & 14.3 & 55.8 & 42.1 & 43.3 & 12.6 & 39.4 & 8.8 & 36.3 & 51.0 & 62.1 & 14.3 & 17.5 & 31.6 & 7.6 & 44.9 & 33.4 & 32.6 \\
$\surd$ & $\surd$ & $\surd$ & $\surd$ & \bf{L} & 62.9 & 55.5 & 43.7 & 14.9 & 13.6 & 57.7 & 52.4 & 50.9 & 13.3 & 45.4 & 4.0 & 30.2 & 55.6 & 67.0 & 3.8 & 23.1 & 39.4 & 5.5 & 50.7 & 29.3 & 35.9 \\
\bottomrule
\end{tabular}
}
\end{table*}

We use the following abbreviations for representing each algorithmic step of the proposed method:
\begin{compactitem}
\item CS: Using contrast scores to mine class-specific object proposals.
\item AS: Using activation scores to mine class-specific object proposals.
\item MIL: Using multiple instance learning to select confident object candidates.
\item Seg: Using segmentation cues to refine the selected object proposals.
\item FT: Using the mined object candidates to fine-tune the detection network.
\end{compactitem}
We evaluate two base CNN models for detector training in our experiments. 
The first one is based on the AlexNet~\cite{krizhevsky2012imagenet} which is referred to as \textbf{S} for a small network. 
The second one is based on the VGGNet model~\cite{simonyan2014very} which is referred to as \textbf{L} for a large network.

We compare the proposed algorithm with state-of-the-art methods for weakly supervised object localization, including the MIL based methods~\cite{siva2012defence,cinbis2017weakly,song2014learning,song2014weakly,bilen2014weakly,bilen2015weakly}, topic model~\cite{shi2013bayesian}, latent category learning~\cite{wang2014weakly}, and recent CNN based models~\cite{bilen2016weakly,kantorov2016contextlocnet,jie2017deep,diba2016weakly}. 
For fair comparisons, we do not include methods that use bounding box annotations for training.

Table~\ref{table:corloc} shows evaluation results in terms of CorLoc on the PASCAL VOC 2007 \emph{trainval} set. 
Our method achieves an average of 59.0\% CorLoc for all 20 categories.
Compared to the MIL based approaches~\cite{siva2012defence,bilen2015weakly,cinbis2017weakly}, we achieve significant improvements by 10 to 20 points.
While these approaches use sophisticated model initialization or optimization strategies for improving MIL, the inevitable noise in the initial collection of category-independent proposals limits the performance of trained object detectors during MIL iterations.
Some of the existing methods rely on hand-crafted features~\cite{shi2013bayesian} or pre-trained CNN features~\cite{wang2014weakly} for representations. 
In contrast, we learn discriminative feature representations progressively for object localization and achieve significant performance improvements (e.g., over 10\% gain compared to~\cite{wang2014weakly}).
Recent CNN based approaches~\cite{bilen2016weakly,kantorov2016contextlocnet,diba2016weakly} also achieve promising results by end-to-end feature representation learning. 
Compared to these methods, our method performs favorably 
with the proposed progressive representation adaptation.

Table~\ref{table:ap} shows the detection average precision (AP) performance on the PASCAL VOC 2007 \emph{test} set. 
With 42.5\% mean average precision (mAP), our full model achieves performs favorably against the state-of-the-art approaches.
Several existing methods~\cite{wang2014weakly,bilen2015weakly,bilen2014weakly,song2014learning,song2014weakly} use pre-trained networks to extract features for object detector learning and do not fine-tune the entire network. 
In contrast, we progressively adapt the network from whole-image classification to object detection. 
This domain adaptation step helps learn better object detectors from weakly annotated data.
Unlike previous work that relies on noisy and class-independent proposals to select object candidates, we mine purer, and class-specific proposals for MIL training, which alleviate the negative effects by background clutter and confusion with similar objects. 

We show in Table~\ref{table:ap-2010} and~\ref{table:ap-2012} the detection average precision performance on the PASCAL VOC 2010 and 2012 datasets, respectively. 
Our method achieves comparable performance with the state-of-the-art WSL methods. 
Similar to the results on VOC 2007 dataset, our algorithm achieves better localization performance for animal and vehicle than that for furniture classes. 
It is difficult to detect indoor objects in a weakly supervised manner due to large appearance variations and background clutter. 
Using a deeper model (e.g., VGGNet), we can learn more effective feature representations for object localization and achieve better performance than that of using the AlexNet.

To validate the effectiveness of the proposed method on a large number of object classes, we evaluate the method using the ILSVRC 2013 detection dataset.
Table~\ref{table:map} shows the mAP performance over 200 classes on the ILSVRC 2013 dataset.\footnote{The results of Wang~\etal~\cite{wang2014weakly} and Diba~\etal~\cite{diba2016weakly} are obtained on the \emph{val} set.}
Our results show that we can obtain an additional 1.3\% of performance gain on this challenging dataset.

\begin{figure*}[t]
\centering
\begin{tabular}{c}
\includegraphics[width=\linewidth]{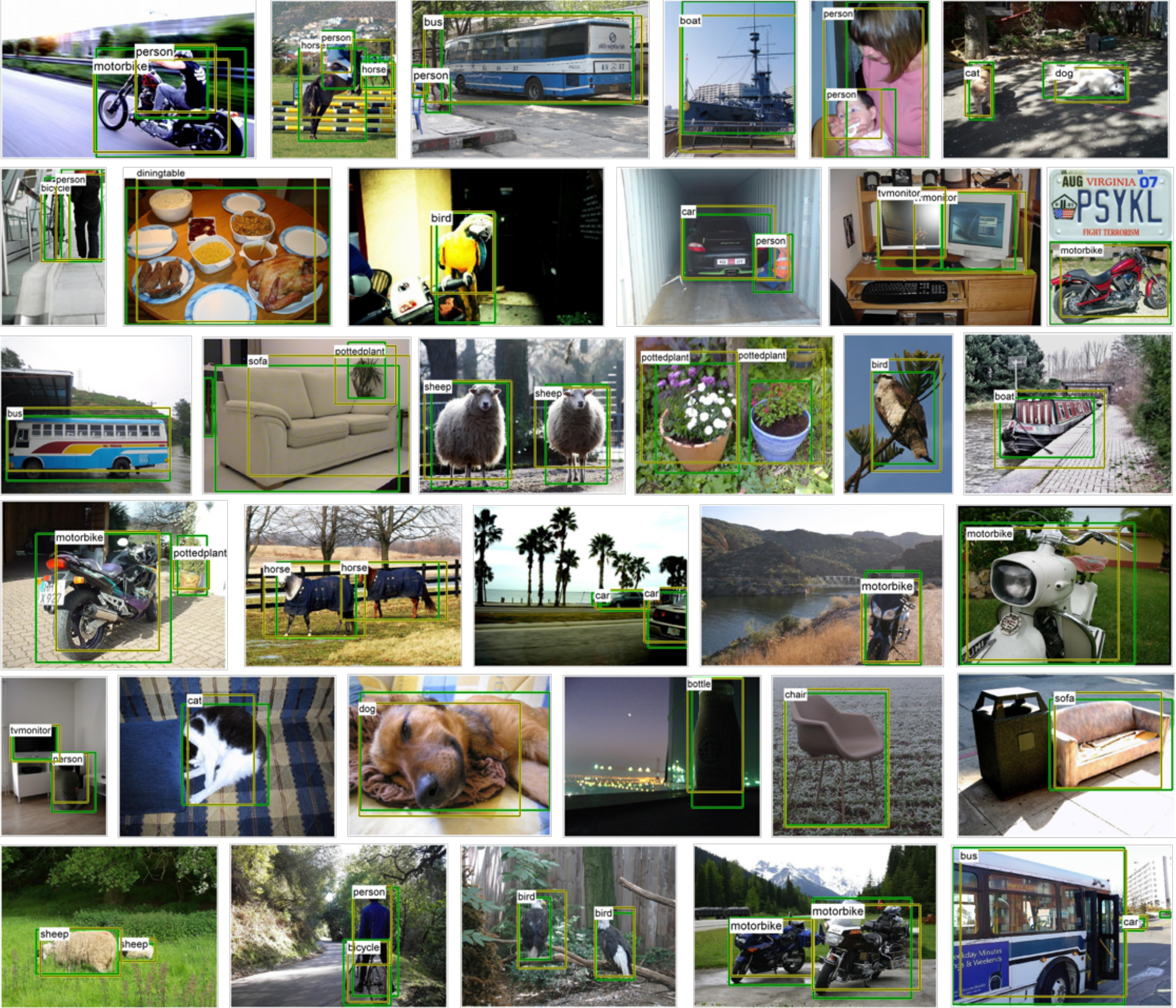} \\
\end{tabular}
\caption{\textbf{Sample detection results on the PASCAL VOC 2007 \emph{test} set}. 
Green boxes indicate ground-truth instance annotation. 
Yellow boxes indicate correction detections (with IoU $\geq 0.5$). 
For all the testing results, we set threshold of detection as 0.8 and use NMS to remove duplicate detections.}
\label{figure:tp}
\end{figure*}

We show sample detection results on the PASCAL VOC 2007 \emph{test} set in Figure~\ref{figure:tp}. 
Our algorithm can detect objects under different scales, lighting conditions, and partial occlusions.

\begin{table}[t]
\caption{\textbf{Object detection on the ILSVRC 2013 dataset.}}
\label{table:map}
\centering
\begin{tabular}{ccccc|c}
\toprule
\multicolumn{5}{l|}{Methods} & mAP \\
\midrule
\multicolumn{5}{l|}{Wang \etal \cite{wang2014weakly}} & 6.0 \\
\multicolumn{5}{l|}{Diba \etal \cite{diba2016weakly}} & 16.3 \\
\midrule
CS & AS & MIL & Seg & FT & \\
$\surd$ & & $\surd$ & & \bf{S} & 7.7 \\
$\surd$ & & $\surd$ & & \bf{L} & 10.8 \\
$\surd$ & $\surd$ & $\surd$ & $\surd$ & \bf{L} & 12.1 \\
\bottomrule
\end{tabular}
\end{table}

\setlength{\tabcolsep}{8pt}
\begin{table}[t]
\caption{\textbf{Different mask-out strategies in terms of average correct localization from top $M$ proposals}.}
\label{table:maskout}
\centering
\resizebox{\linewidth}{!}{
\begin{tabular}{c|cccc}
\toprule
Mask-out strategy & $M$=1 & $M$=10 & $M$=50 & $M$=100 \\
\midrule
\emph{In-Out} & 31.8 & 73.8 & 82.9 & 84.2 \\
\emph{Whole-Out} & 29.6 & 64.9 & 76.0 & 78.5 \\
\emph{In} & 32.7 & 71.0 & 79.9 & 81.8 \\
\bottomrule
\end{tabular}
}
\end{table}

\begin{figure}[t]
\centering
\includegraphics[width=0.8\linewidth]{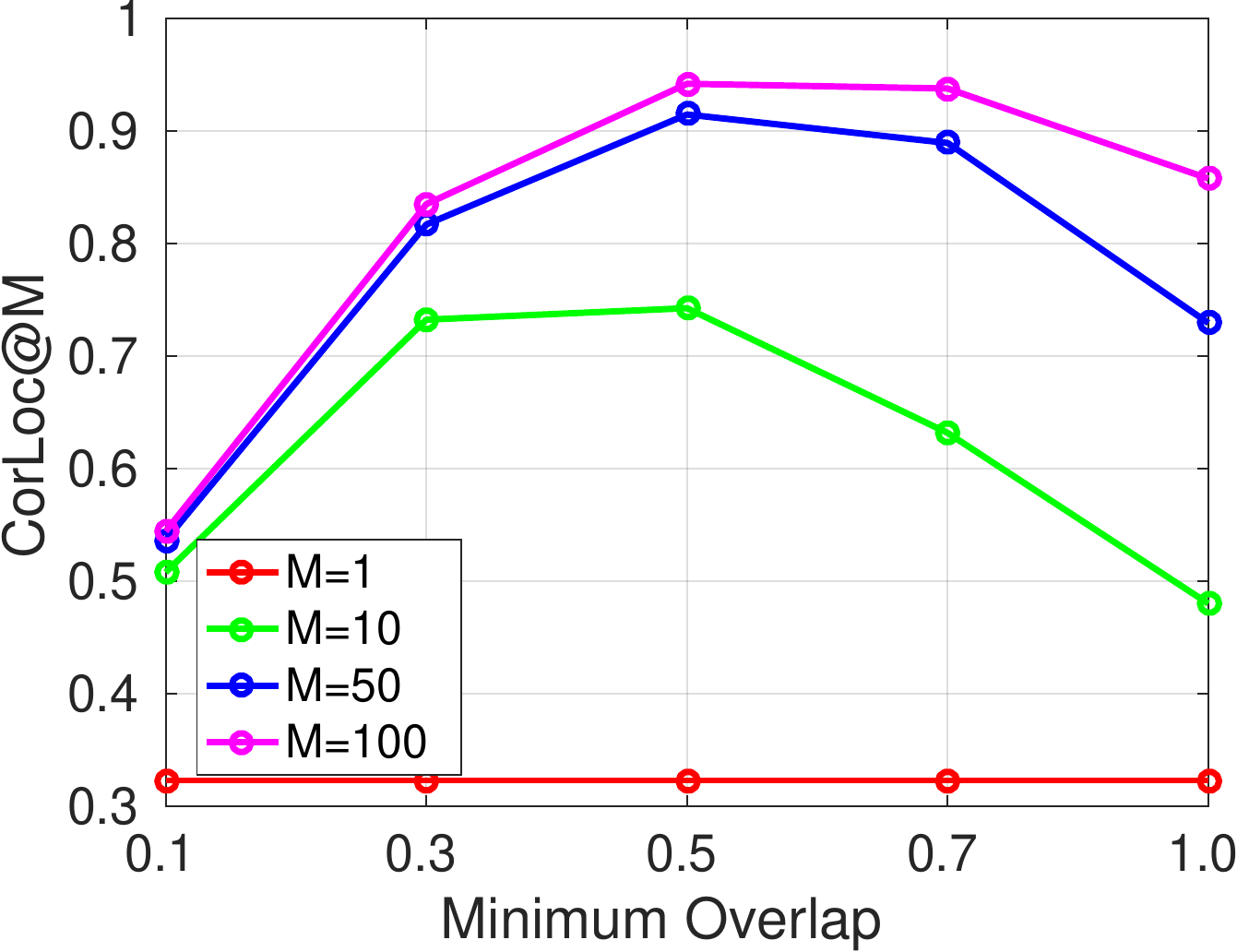} \\ 
\includegraphics[width=0.8\linewidth]{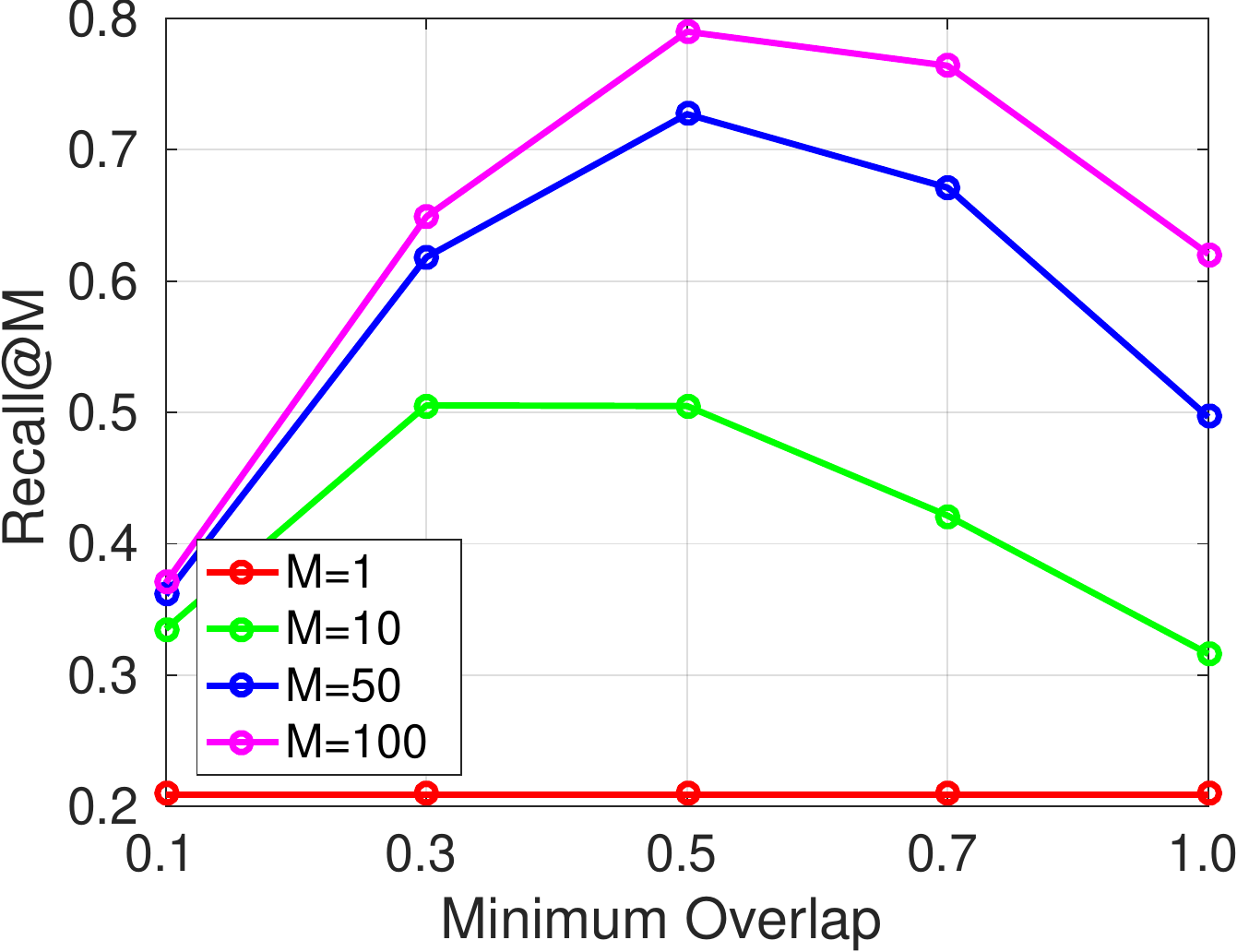} \\
\caption{\textbf{Effect of the minimum bounding box overlap for mining class-specific proposals}. }
\label{figure:iou}
\end{figure}

\subsection{Ablation study}
\label{section:ablation}

\begin{figure*}[!t]
\centering
\begin{tabular}{@{}cccc@{}}
\includegraphics[width=0.23\linewidth]{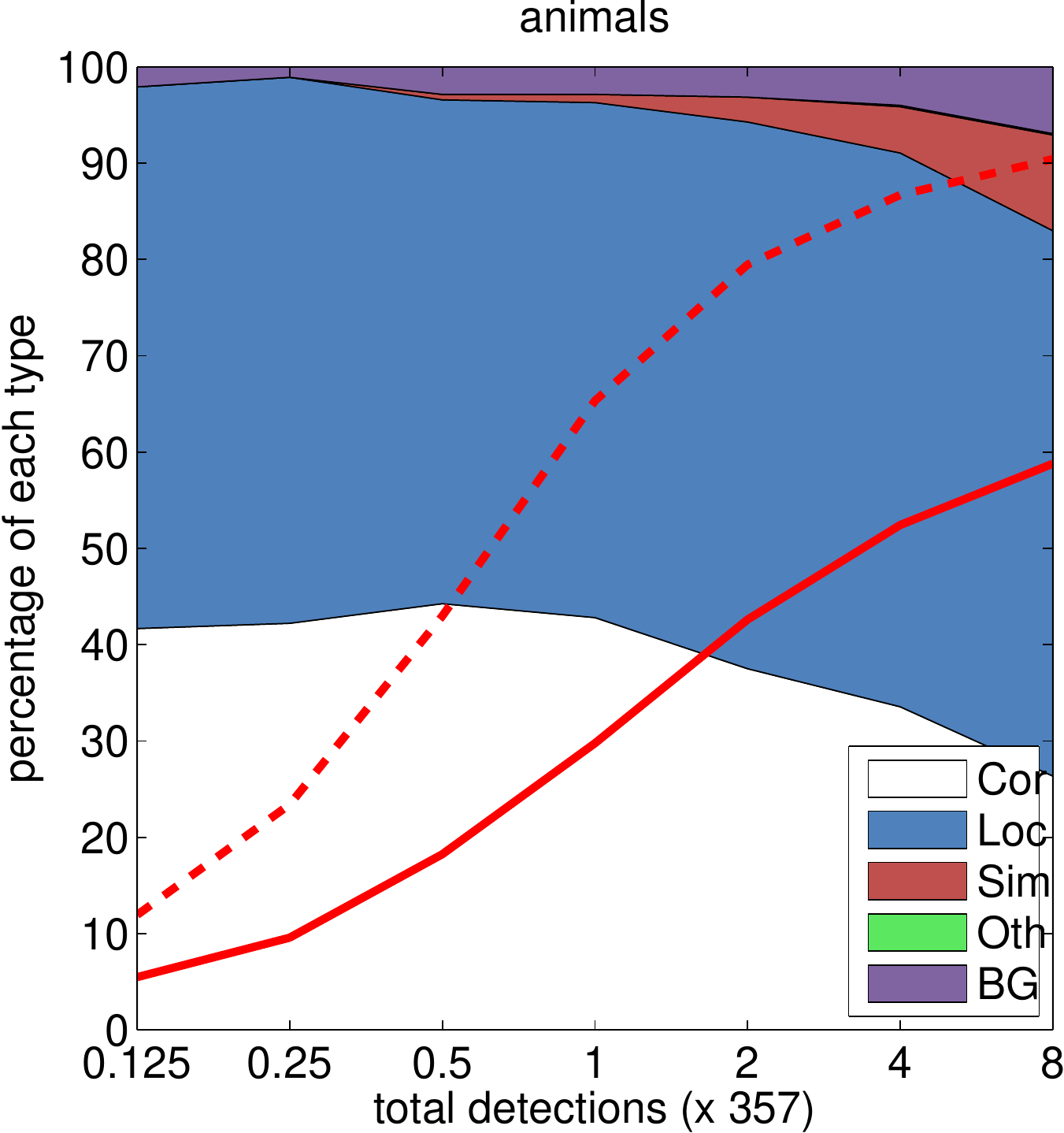} & \includegraphics[width=0.23\linewidth]{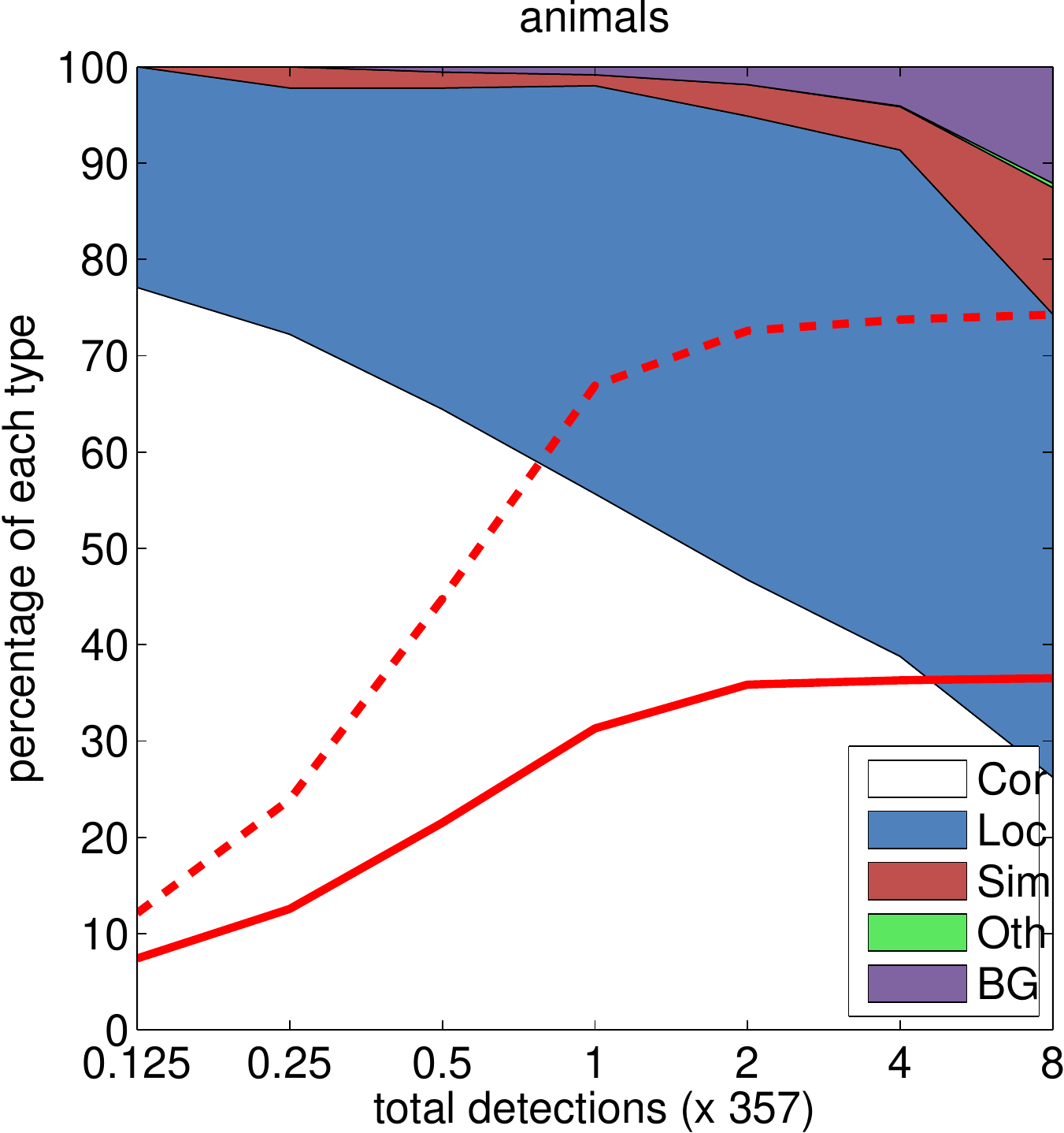} & \includegraphics[width=0.23\linewidth]{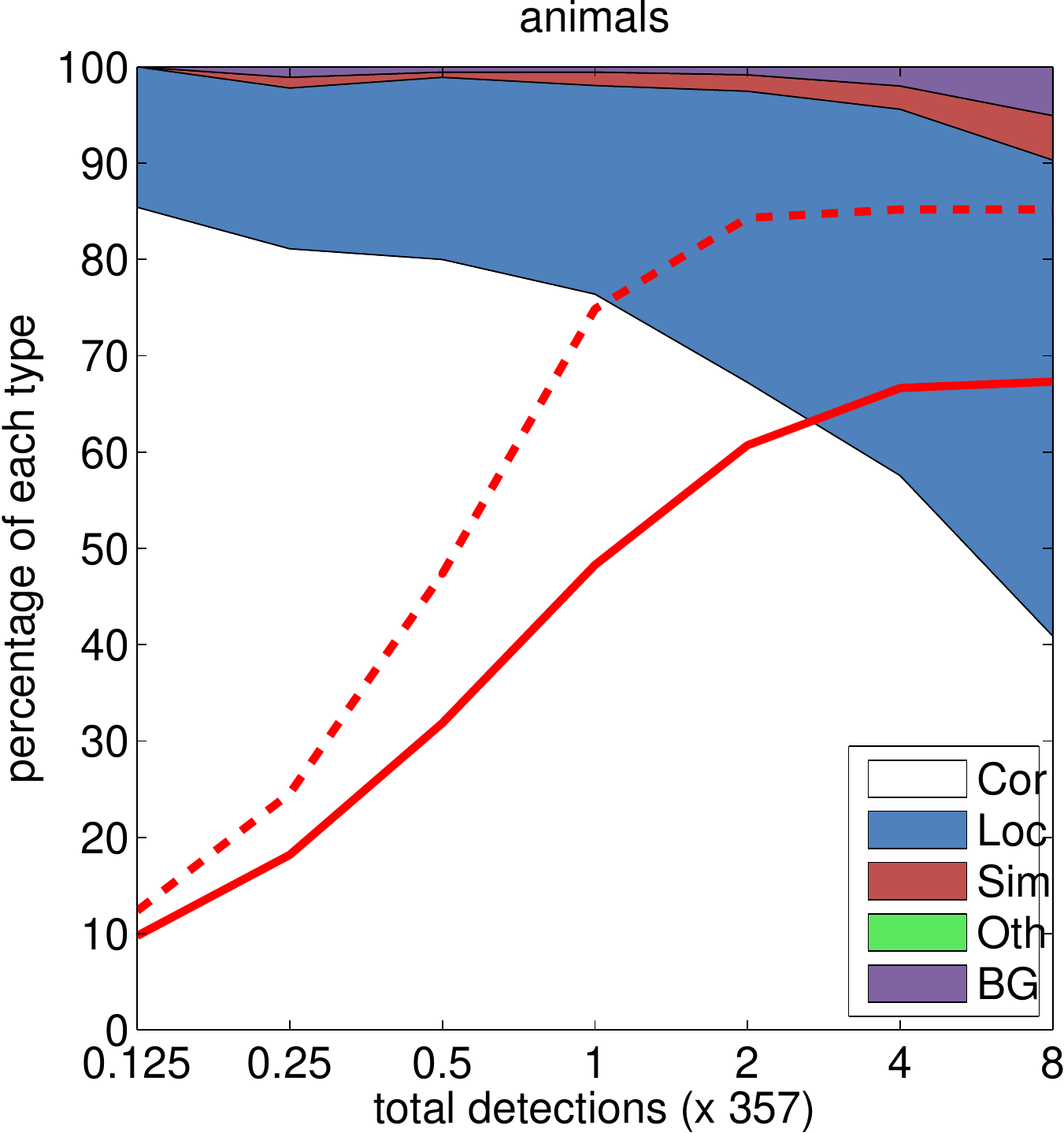} & \includegraphics[width=0.23\linewidth]{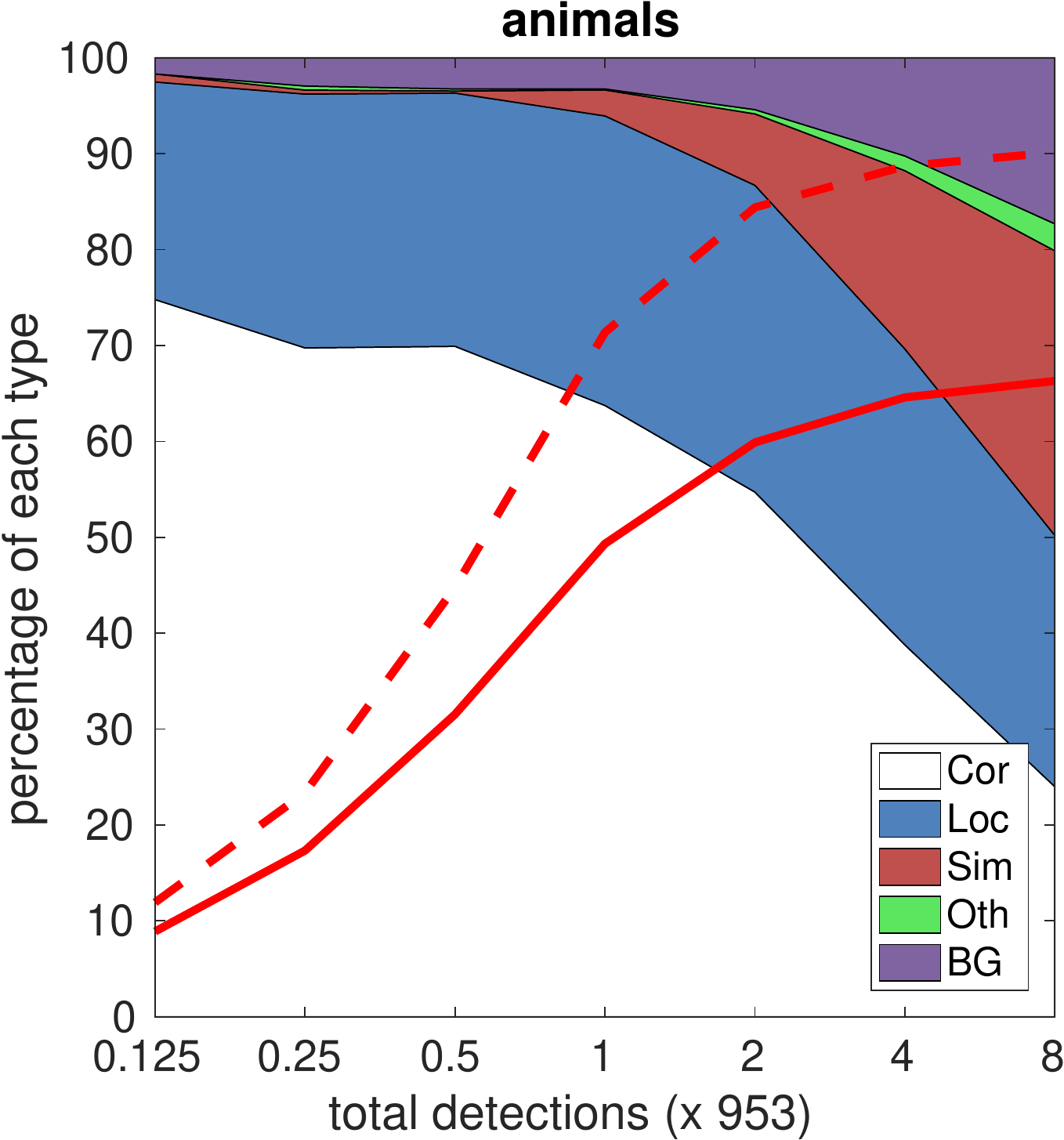} \\
\includegraphics[width=0.23\linewidth]{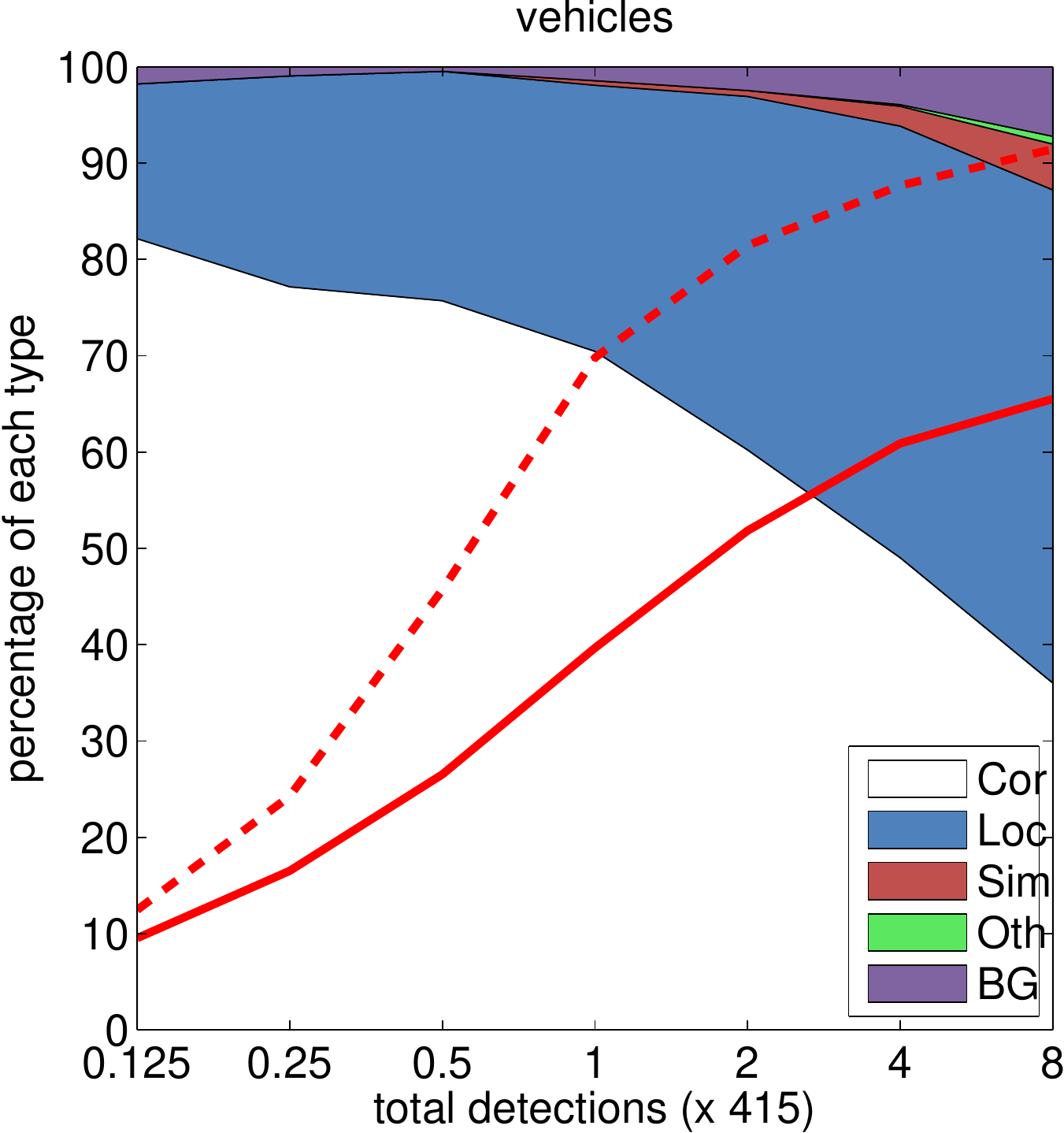} & \includegraphics[width=0.23\linewidth]{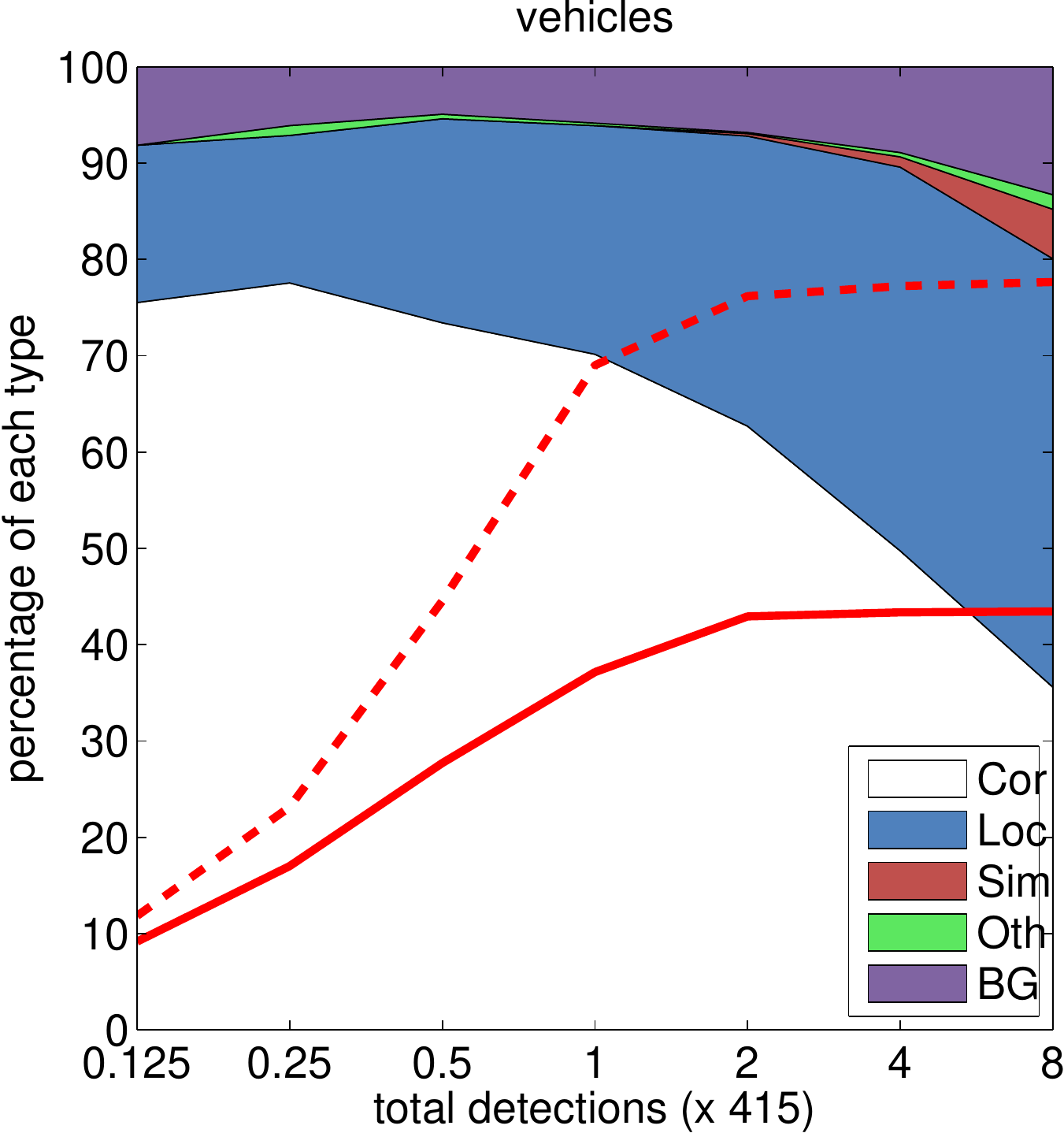} & \includegraphics[width=0.23\linewidth]{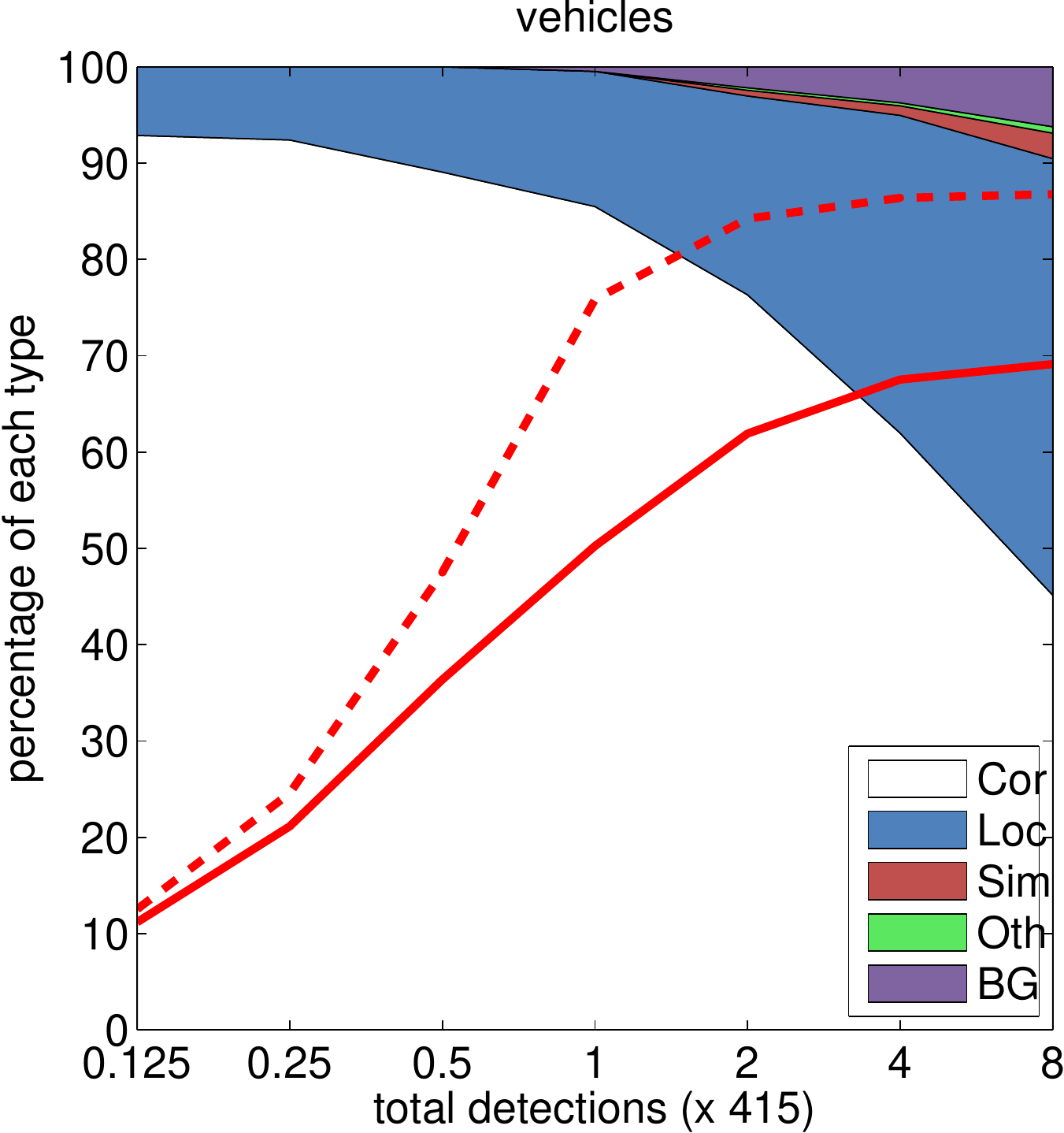} & \includegraphics[width=0.23\linewidth]{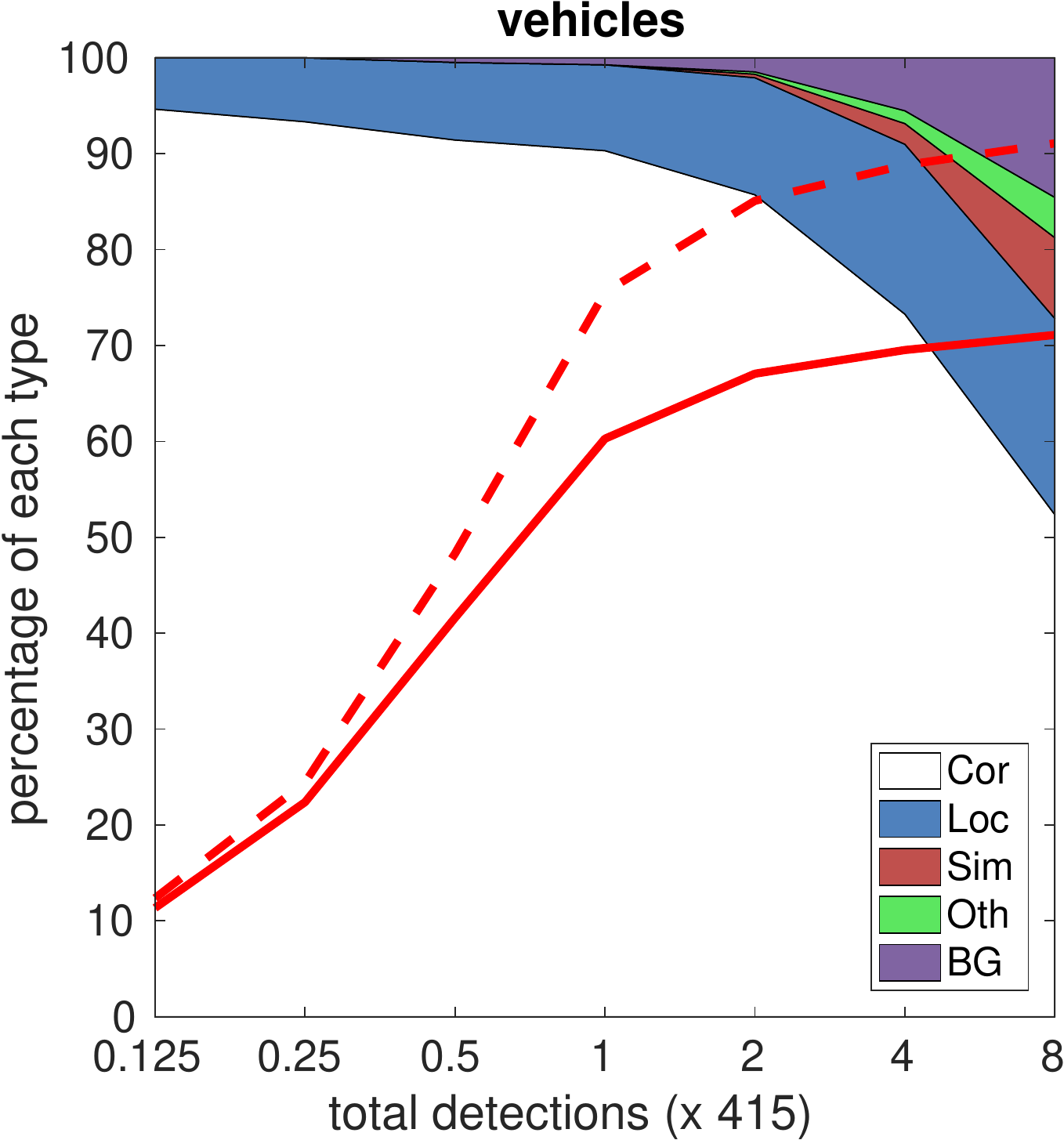} \\
\includegraphics[width=0.23\linewidth]{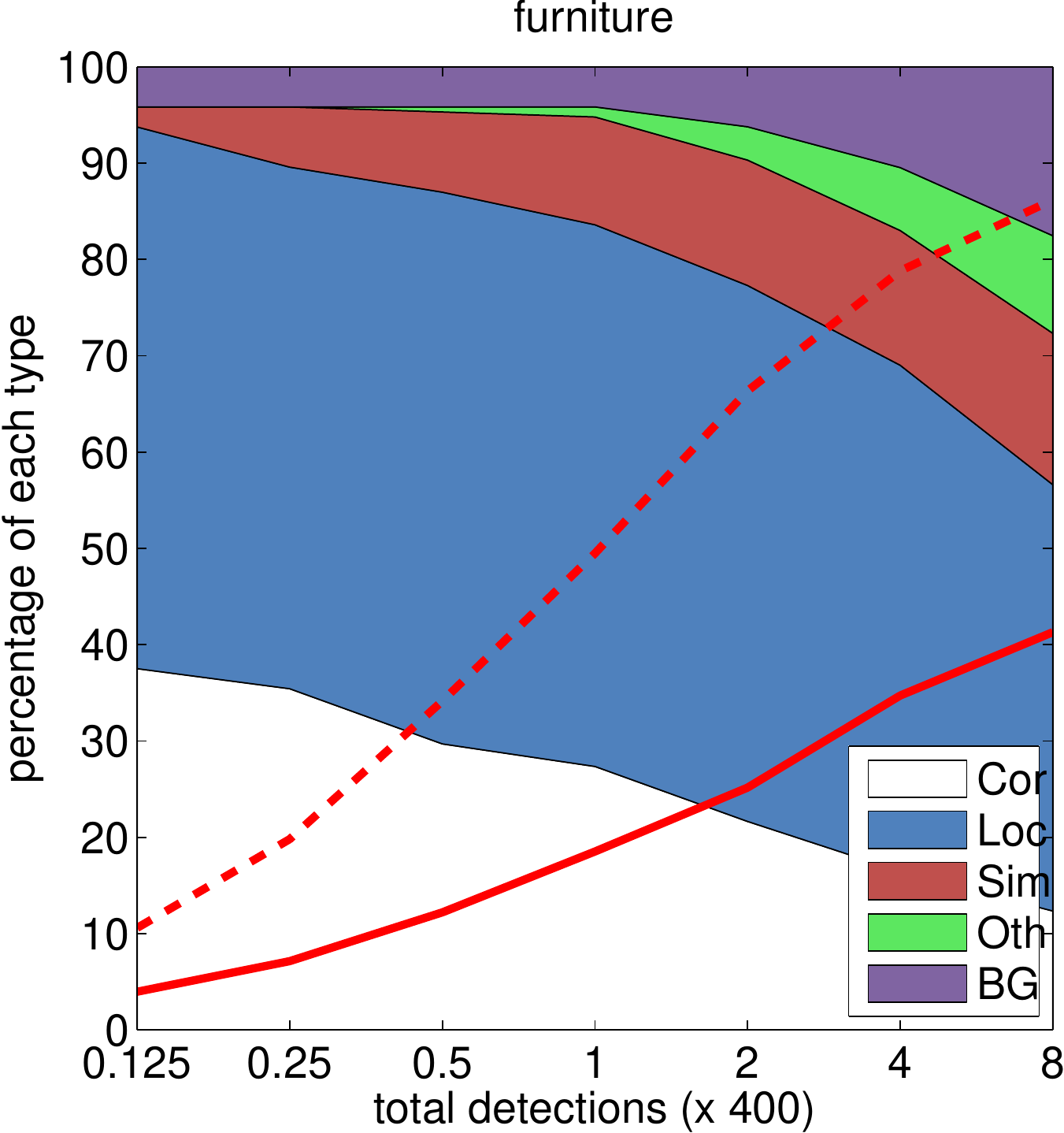} & \includegraphics[width=0.23\linewidth]{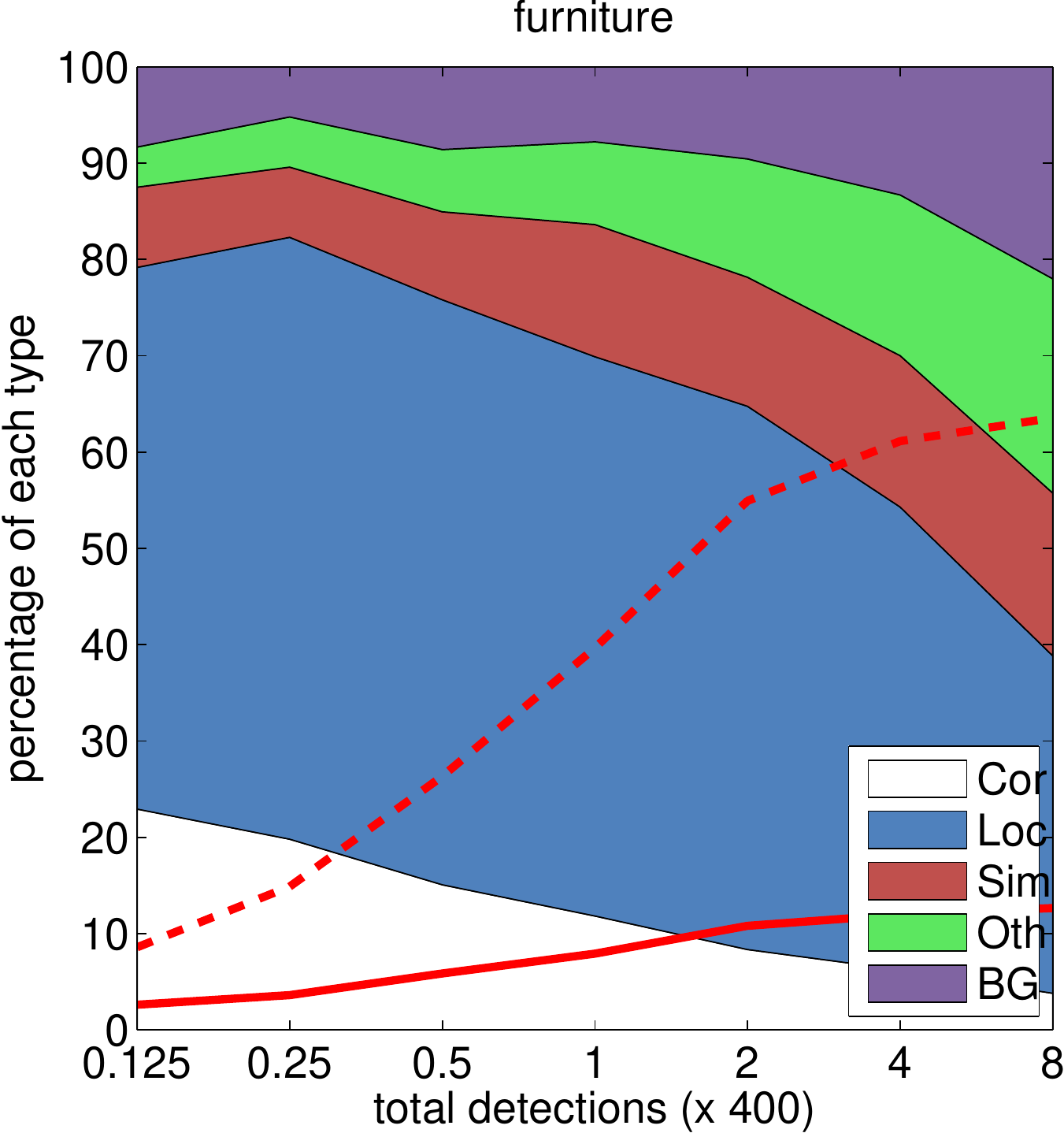} & \includegraphics[width=0.23\linewidth]{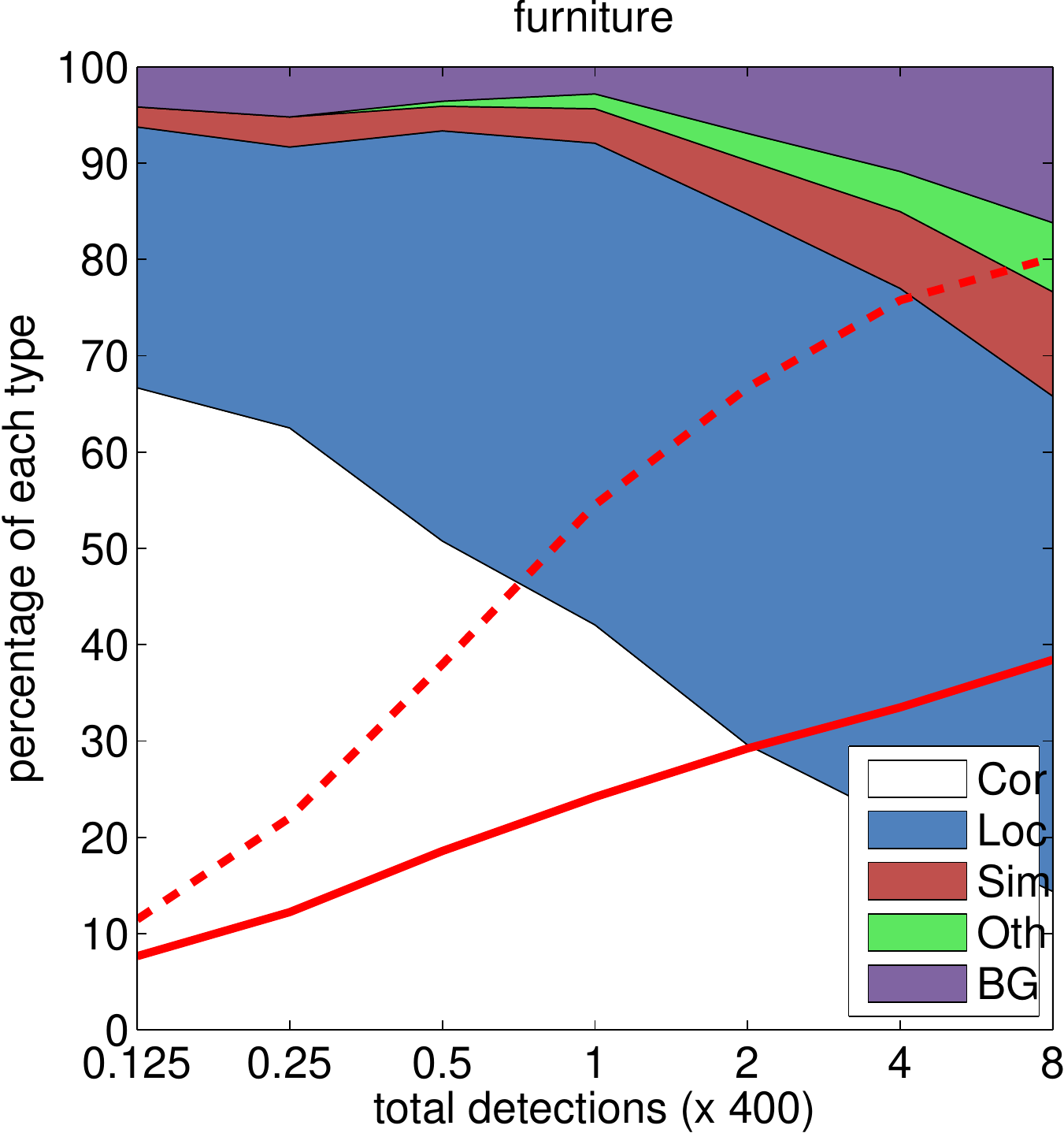} & \includegraphics[width=0.23\linewidth]{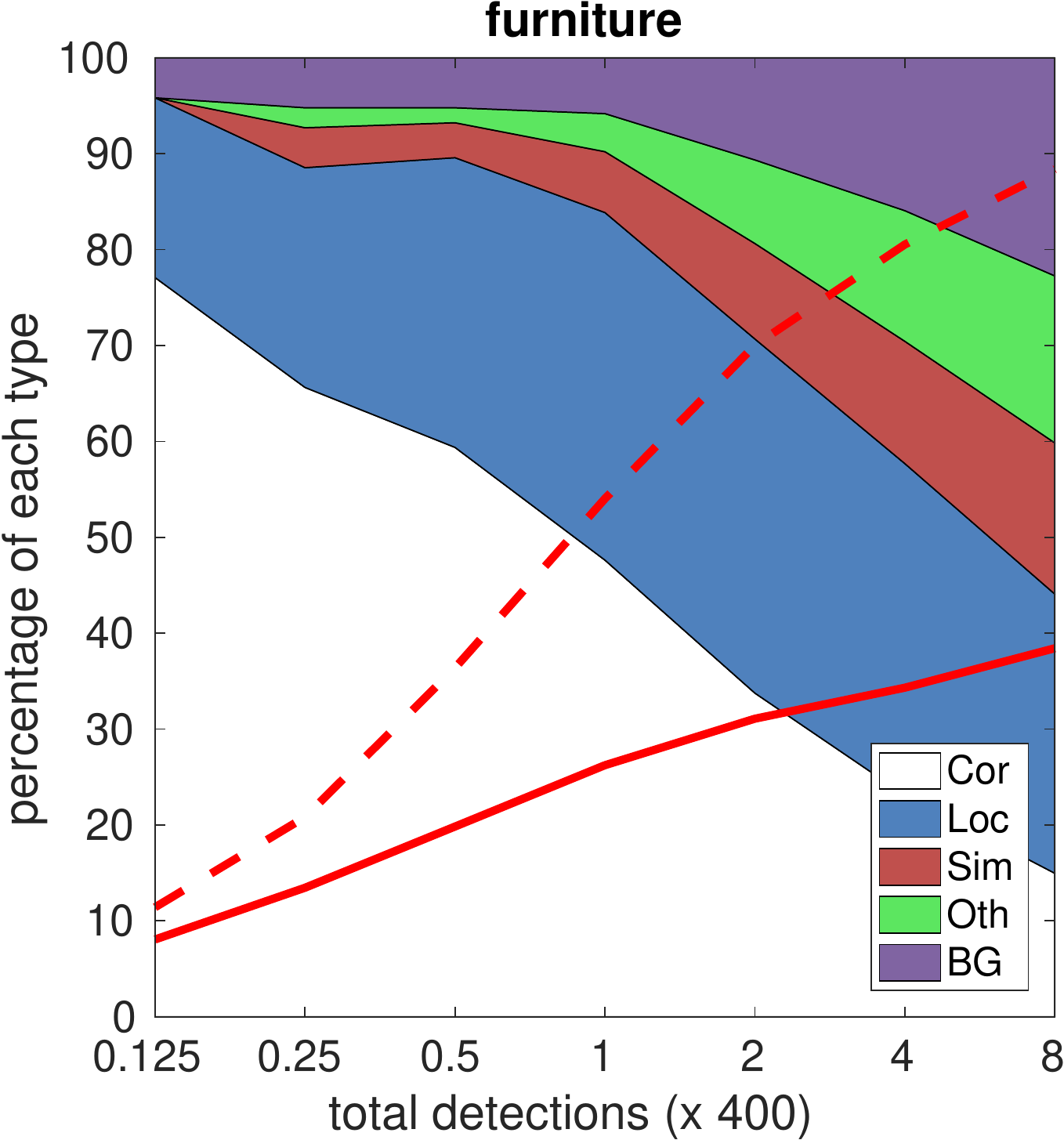} \\
CS+MIL & MIL+FT & CS+MIL+FT & CS+AS+MIL+Seg+FT \\
\end{tabular}
\caption{\textbf{Detector error analysis}. The detections are categorized into five types of correct detection (Cor), false positives due to poor localization (Loc), confusion with similar objects (Sim), confusion with other VOC objects (Oth), and confusion with background (BG). Each plot shows types of detection as top detections increase. Line plots show recall as function of the number of objects by IoU $\geq 0.5$ (solid) and IoU $\geq 0.1$ (dash). The VGGNet is used as the base network for training object detectors.}
\label{figure:analysis}
\end{figure*}

\begin{figure}[t]
\centering
\includegraphics[width=\linewidth]{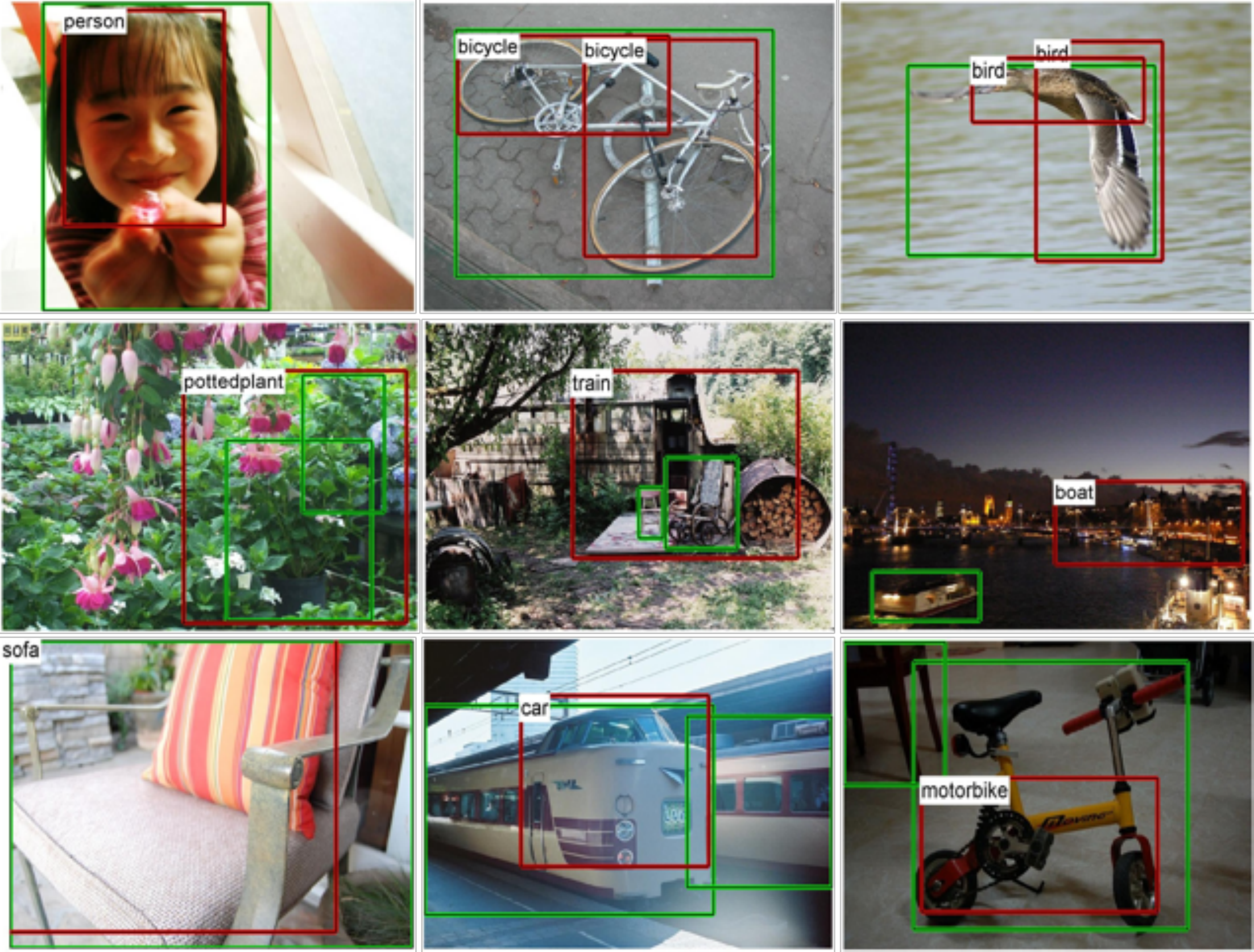}
\caption{\textbf{Sample results of detection errors}. Green boxes indicate ground-truth instance annotation. Red boxes indicate false positives.}
\label{figure:fp}
\end{figure}

To analyze the contribution of each component in our method, we examine the performance of the proposed algorithm using different configurations.
The last grouped rows of Table~\ref{table:corloc} show how these variants perform in terms of CorLoc on the PASCAL VOC 2007 \emph{trainval} dataset.
We achieve an average CorLoc of 31.8\% by directly using the top-ranked class-specific object proposals.
Using MIL to select confident objects, we obtain 41.2\% with about 10 points improvement.
The results demonstrate that MIL iterations significantly help to select better object proposals. 
The performance boost comes from: (1) the mined object proposals are less noisy and can discard background clutters, and (2) the mined object proposals are class-specific and can alleviate potential confusion with similar objects. 
Furthermore, adding the detection network fine-tuning step, we obtain an average CorLoc of 
49.8\% using the AlexNet and 52.4\% using the deeper VGGNet~\cite{simonyan2014very}. 
Such network fine-tuning further boosts the performance by another 10 points as this step helps learn adapted feature representations for object localization. 
In addition, we obtain a 1.9\% gain with activation scores and obtain another 4.5\% gain when using segmentation cues for object bounding box refinement. 
Our results validate that improving the quality of the selected object candidates can further boost the performance of object localization.

The last grouped rows of Table~\ref{table:ap} show our detection AP performance on the PASCAL VOC 2007 \emph{test} set. 
We use the method by Song~\etal~\cite{song2014learning} as our MIL baseline.
A straightforward approach to train detector uses proposals selected by MIL.
However, this simple combination of MIL and detector training only gives marginal performance improvement from 22.7\% to 23.0\% because the selected proposals by MIL are too noisy 
to train object detection network effectively. 
Using the top-ranked object proposals based on the adapted classification network, we achieve significant improvement from 23.0\% to 31.0\%, highlighting the importance of progressive adaptation.
Using the deeper VGGNet~\cite{simonyan2014very}, we can achieve a large improvement from 26.2\% to 39.5\%. 
We also evaluate the performance using the best proposal ($M=1$) mined by the mask-out strategy for detection adaptation. 
This method achieves 19.5\% mAP using AlexNet and 20.5\% using the VGGNet. 
Without the MIL step, the results are inferior due to noisy training samples. 
Combined with contrast and activation scores, we improve the mAP performance by 1.7\%. 
The results show that fusion of different levels of feature representations helps mine better object proposals (with higher recall). 
With proposal refinement, we can further improve the performance from 41.2\% to 42.5\%.
These experimental results validate the importance of the progressive adaptation steps proposed in this work.

\subsection{Parameter analysis}
\label{section:parameter}

We present detailed parameter analysis to study the effect of hyper-parameters on the PASCAL VOC 2007 dataset, including the minimum bounding box overlap $t$ for mining class-specific proposals and the number of selected proposals $M$ by MIL.
We also evaluate different mask-out strategies when computing contrast scores. 
For parameter analysis experiments, we use the \emph{train} set for training and the \emph{val} set for validation. 

Figure~\ref{figure:iou} shows the effect of the minimum overlap $t$ for mining class-specific proposals. 
Similar to the CorLoc metric, we compute CorLoc@M as the percentage of positive images where at least one of top $M$ proposals has an IoU overlap with a ground-truth bounding box greater than $t$. 
When $M=1$, this metric falls back to CorLoc. 
We also compute the recall at top $M$ proposals. 
According to the validation results, we set $t=0.5$ for mining class-specific proposals.

Table~\ref{table:maskout} shows the CorLoc@M results using different mask-out strategies. 
The \emph{In-Out} strategy computes the classification score difference between the selected object proposal and its mask-out image. 
The \emph{Whole-Out} computes the score difference between the whole image and a mask-out image~\cite{Bazzani:WACV16}. 
The \emph{In} strategy directly selects top object proposals using the classification scores computed directly from the proposals. 
The results show that the proposed \emph{In-Out} strategy consistently outperforms \emph{Whole-Out} for different $M$. 
Only using classification score of the proposal itself can also collect good proposals because our classification adaptation step trains the network to be sensitive to object categories of the target datasets.
As the classification network is fine-tuned using the whole image, the mask-out image can provide additional discriminative power for ranking object proposals. 

Figure~\ref{figure:iou} and Table~\ref{table:maskout} show that the fraction of correct localization/recall increases as the amount of mined proposals $M$ increases. 
We note that more noise (proposals from other categories or background clutter) will also be introduced as $M$ increases. A large number of $M$ also increases the computational cost for the subsequent MIL step. 
We strike a balance and set $M=50$ throughout the experiments. 

\subsection{Error analysis}
\label{section:error}

In Figure~\ref{figure:analysis}, we apply the detector error analysis tool from Hoiem~\etal~\cite{hoiem2012diagnosing} to analyze errors of our detector.
Comparing the first and third columns, we achieve significant improvement of localization performance by detection adaptation. 
Fine-tuning the network for object-level detection helps learn discriminative appearance model for object categories, particularly for animals and furniture classes.
The comparison between the second and third columns highlights the importance of class-specific proposal mining step.
We attribute the performance boost to the classification adaptation that fine-tunes the network from 1000-way single-label classification (source) to 20-way multi-label classification task (target).
The last column shows our full model. 
With more accurate object proposal candidates from progressive adaptation steps, we can further reduce the localization errors.

These detector analysis plots also show that the majority of errors come from \emph{inaccurate localization}. 
We show a few sample results in Figure~\ref{figure:fp}. 
Our model often detects the correct category of an object instance but fails to predict a sufficiently tight bounding box, e.g., IoU $\in [0.1,0.5)$.
Typical errors with imprecise localization include detecting a human face, a bicycle wheel and a bird body, as shown in the first rows of Figure~\ref{figure:fp}. 
Sometimes the detector gets confused with background clutter or semantically similar objects. 
The last two rows of Figure~\ref{figure:fp} show the detection errors due to confusion with background and similar objects, respectively. 
For example, we detect plant in the lake and claim to detect potted plants, and incorrectly detect a chair as a sofa. 
The detection analysis suggests that the learned model makes sensible errors. 
We believe that we can further improve the localization performance of our model by incorporating techniques for addressing the inaccurate localization issues~\cite{Dai-CVPR-2012,zhang2015improving}.

\section{Conclusions}
\label{section:conclusion}

The weakly supervised setting is of great importance for large-scale practical applications as it does not require intensive and expensive instance-level labeling work. 
In this paper, we present a progressive representation adaptation approach to tackle the weakly supervised object localization problem. 
In classification adaptation, we transfer the classifiers from source to target domains using a multi-label loss function for training a multi-label classification network. 
In detection adaptation, we transfer adapted classifiers to object detectors. 
We extensively evaluate the proposed progressive representation adaptation algorithm on the PASCAL VOC and ILSVRC datasets and achieve favorable results against the state-of-the-art methods. 

\bibliographystyle{IEEEtran}
\bibliography{mybib}

\vspace{-8mm}
\begin{IEEEbiography}[{\includegraphics[width=1in,height=1.25in,clip,keepaspectratio]{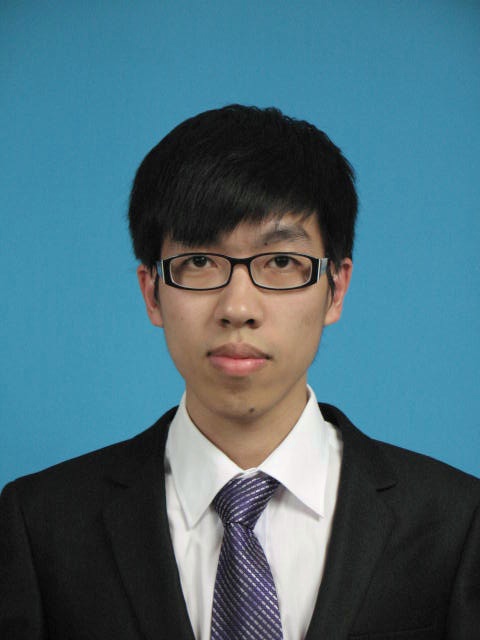}}]{Dong Li} received the Ph.D. degree in Electronic Engineering from Tsinghua University in 2017. He also received the B.E. degree in Electronic Engineering from Tsinghua University in 2012. His research interests include computer vision, machine learning and particularly deep learning for visual recognition.
\end{IEEEbiography}

\vspace{-10mm}
\begin{IEEEbiography}[{\includegraphics[width=1in,height=1.25in,clip,keepaspectratio]{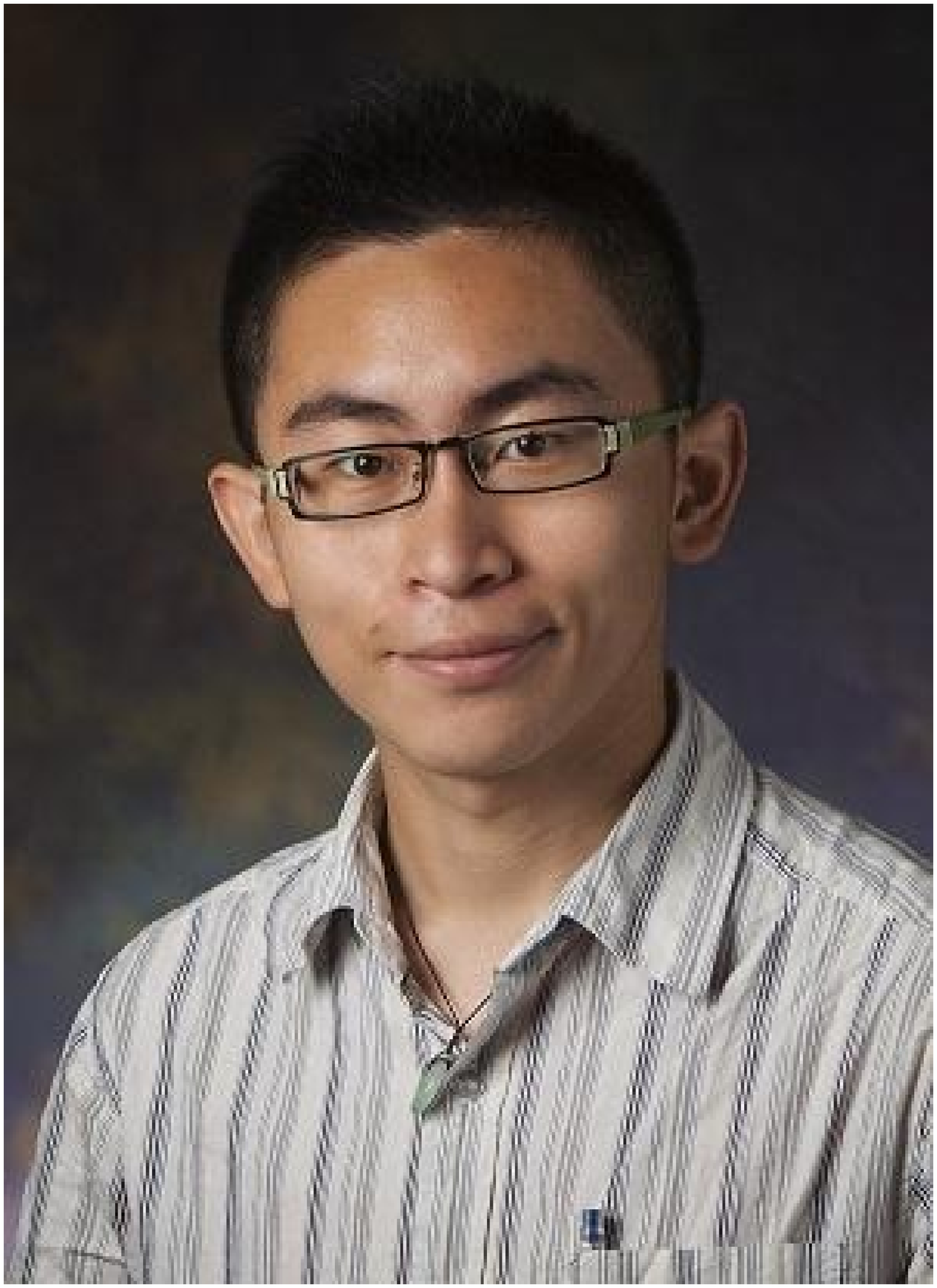}}]{Jia-Bin Huang} 
is an assistant professor in the Bradley Department of Electrical and Computer Engineering at Virginia Tech. He received the B.S. degree in Electronics Engineering from National Chiao-Tung University, Hsinchu, Taiwan and his Ph.D. degree in the Department of Electrical and Computer Engineering at University of Illinois, Urbana-Champaign in 2016. He is a member of the IEEE.
\end{IEEEbiography}

\vspace{-10mm}
\begin{IEEEbiography}[{\includegraphics[width=1in,height=1.25in,clip,keepaspectratio]{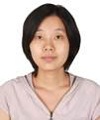}}]{Yali Li} received the B.E. degree with Excellent Graduates Award from Nanjing University, China, in 2007 and the Ph.D degree from Tsinghua University, Beijing, China, in 2013. Currently she is an assistant professor in the Department of Electronic Engineering at Tsinghua University. Her research interests include computer vision, visual recognition, object detection and video understanding, etc.
\end{IEEEbiography}

\vspace{-10mm}
\begin{IEEEbiography}[{\includegraphics[width=1in,height=1.25in,clip,keepaspectratio]{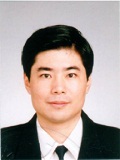}}]{Shengjin Wang} received the B.E. degree from Tsinghua University, China, in 1985 and the Ph.D. degree from the Tokyo Institute of Technology, Tokyo, Japan, in 1997. From 1997 to 2003, he was a member of Research Staff in the Internet System Research Laboratories, NEC Corporation, Japan. Since 2003, he has been a Professor with the Department of Electronic Engineering, Tsinghua University. He has published over 80 papers on image processing, computer vision, and pattern recognition. He is the holder of ten patents. His current research interests include image processing, computer vision, video surveillance, and pattern recognition. He is a member of the IEEE and the IEICE.
\end{IEEEbiography}

\vspace{-10mm}
\begin{IEEEbiography}[{\includegraphics[width=1in,height=1.25in,clip,keepaspectratio]{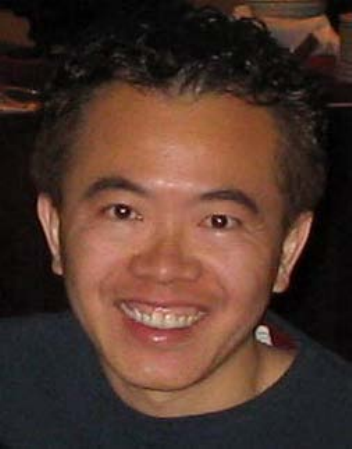}}]{Ming-Hsuan Yang} is a professor in Electrical Engineering and Computer Science at University of California, Merced. He received the Ph.D. degree in computer science from the University of Illinois at Urbana-Champaign in 2000. Prior to joining UC Merced in 2008, he was a senior research scientist at the Honda Research Institute. He served as an associate editor of the IEEE Transactions on Pattern Analysis and Machine Intelligence from 2007 to 2011, and is an associate editor of the International Journal of Computer Vision, Computer Vision and Image Understanding, Image and Vision Computing and Journal of Artificial Intelligence Research. He received the NSF CAREER award in 2012, and the Google Faculty Award in 2009. He is a senior member of the IEEE and the ACM.
\end{IEEEbiography}

\end{document}